\pdfoutput=1

\documentclass[11pt]{article}

\usepackage{acl}

\usepackage{times}
\usepackage{latexsym}

\usepackage[T1]{fontenc}

\usepackage[utf8]{inputenc}

\usepackage{microtype}

\usepackage{inconsolata}

\usepackage{url}
\usepackage{multirow}
\usepackage{multicol}
\usepackage{graphicx}
\usepackage{booktabs}
\usepackage{subfigure}
\usepackage{latexsym}

\usepackage[
            ]{hyperref}
\usepackage{url}
\usepackage{wrapfig}

\usepackage{inconsolata}
\usepackage{eurosym}
\usepackage{todonotes}
\usepackage{subcaption}
\usepackage{amssymb}
\usepackage{amsmath}
\usepackage{enumitem}
\usepackage{array}
\usepackage{longtable}
\usepackage{makecell}
\usepackage{xspace}
\usepackage{xcolor,colortbl}
\usepackage{tcolorbox}
\usepackage{arydshln}
\usepackage{adjustbox}

\usepackage{tikz}
\usepackage[tikz]{bclogo}
\usepackage{pgfplots}
\pgfplotsset{width=1.0\columnwidth}

\usepackage{makecell}
\usepackage{pifont}
\usepackage{bbm}
\usepackage{rotating}
\usepackage{tablefootnote}
\usepackage{soul}
\usepackage{tikz}
\usepackage[tikz]{bclogo}
\usepackage{pgfplotstable}
\usepackage{mdframed}
\usepackage{nameref}
\usetikzlibrary{pgfplots.statistics}

\usepackage{listings}
\usepackage{xcolor}

\definecolor{eclipseStrings}{RGB}{42,0.0,255}
\definecolor{eclipseKeywords}{RGB}{127,0,85}
\colorlet{numb}{magenta!60!black}

\lstdefinelanguage{json}{
    basicstyle=\normalfont\ttfamily,
    commentstyle=\color{eclipseStrings}, 
    stringstyle=\color{eclipseKeywords}, 
    numbers=left,
    numberstyle=\scriptsize,
    stepnumber=1,
    numbersep=6pt,
    showstringspaces=false,
    breaklines=true,
    frame=lines,
    string=[s]{"}{"},
    comment=[l]{:\ "},
    morecomment=[l]{:"},
    literate=
        *{0}{{{\color{numb}0}}}{1}
         {1}{{{\color{numb}1}}}{1}
         {2}{{{\color{numb}2}}}{1}
         {3}{{{\color{numb}3}}}{1}
         {4}{{{\color{numb}4}}}{1}
         {5}{{{\color{numb}5}}}{1}
         {6}{{{\color{numb}6}}}{1}
         {7}{{{\color{numb}7}}}{1}
         {8}{{{\color{numb}8}}}{1}
         {9}{{{\color{numb}9}}}{1}
}

\newmdenv[
  backgroundcolor=blue!05,
  linecolor=quoteborder,
  skipabove=1em,
  skipbelow=0em,
  leftline=true,
  topline=false,
  bottomline=false,
  rightline=false,
  linecolor=blue!66,
  linewidth=4pt
]{githubquote}

\newcommand{\blue}[1]{\textcolor{blue}{#1}}
\newcommand{\brown}[1]{\textcolor{brown}{#1}}

\newcommand{\orange}[1]{\textcolor{orange}{#1}}
\newcommand{\jsonvalue}[1]{\textcolor{black}{#1}}
\newcommand{\jsonkey}[1]{\textcolor{eclipseKeywords}{#1}}

\newcommand{\myslash}{\textbackslash n}

\title{JsonTuning: Towards Generalizable, Robust, and Controllable Instruction Tuning}

\author{Chang Gao$^{\clubsuit}$, Wenxuan Zhang$^\spadesuit$\thanks{Wenxuan Zhang is the corresponding author.}, Guizhen Chen$^\heartsuit$$^\diamondsuit$, Wai Lam$^{\clubsuit}$ \\
$^\clubsuit$The Chinese University of Hong Kong \ $^\spadesuit$Singapore University of Technology and Design\\ $^\heartsuit$DAMO Academy, Alibaba Group \ $^\diamondsuit$Nanyang Technological University}

\begin{document}
\maketitle
\begin{abstract}
Instruction tuning is vital for enhancing the performance of large language models (LLMs), but existing text-to-text methods, referred to as TextTuning, struggle with issues such as generalization, robustness, and controllability due to their lack of explicit task structures. We introduce JsonTuning, a structure-to-structure approach that uses JSON structures to represent tasks. This method improves generalization by clarifying task elements and their relations, boosts robustness by minimizing ambiguity, and enhances controllability by allowing precise control over outputs.
We conduct an extensive comparative analysis between JsonTuning and TextTuning using various language models and benchmarks. Our findings reveal that JsonTuning consistently surpasses TextTuning in terms of performance, robustness, and controllability across different scenarios. By overcoming the limitations of TextTuning, JsonTuning demonstrates significant potential for developing more effective and reliable LLMs capable of handling diverse scenarios\footnote{The code is available at \url{https://github.com/gao-xiao-bai/JsonTuning}.}. 
\end{abstract}

\section{Introduction}
\label{sec:introduction}

The field of natural language processing has witnessed significant advancements driven by large language models (LLMs) such as GPT-3 \citep{GPT-3}, PaLM \citep{PaLM}, and LLaMA \citep{LLaMA}, which excel in various tasks such as machine translation and sentiment analysis. 
However, effectively interpreting and responding to human instructions remains a challenge. Instruction tuning \citep{wei2022finetuned} addresses this challenge by fine-tuning LLMs with explicit task instructions, thereby enhancing their understanding and execution of tasks. This approach has paved the way for the success of instruction-following LLMs like InstructGPT \citep{InstructGPT}, ChatGPT \citep{Chatgpt}, and Claude \citep{Claude3} in a wide range of applications.

 \begin{figure*}[t]
    \centering
    \includegraphics[width=0.86\linewidth]{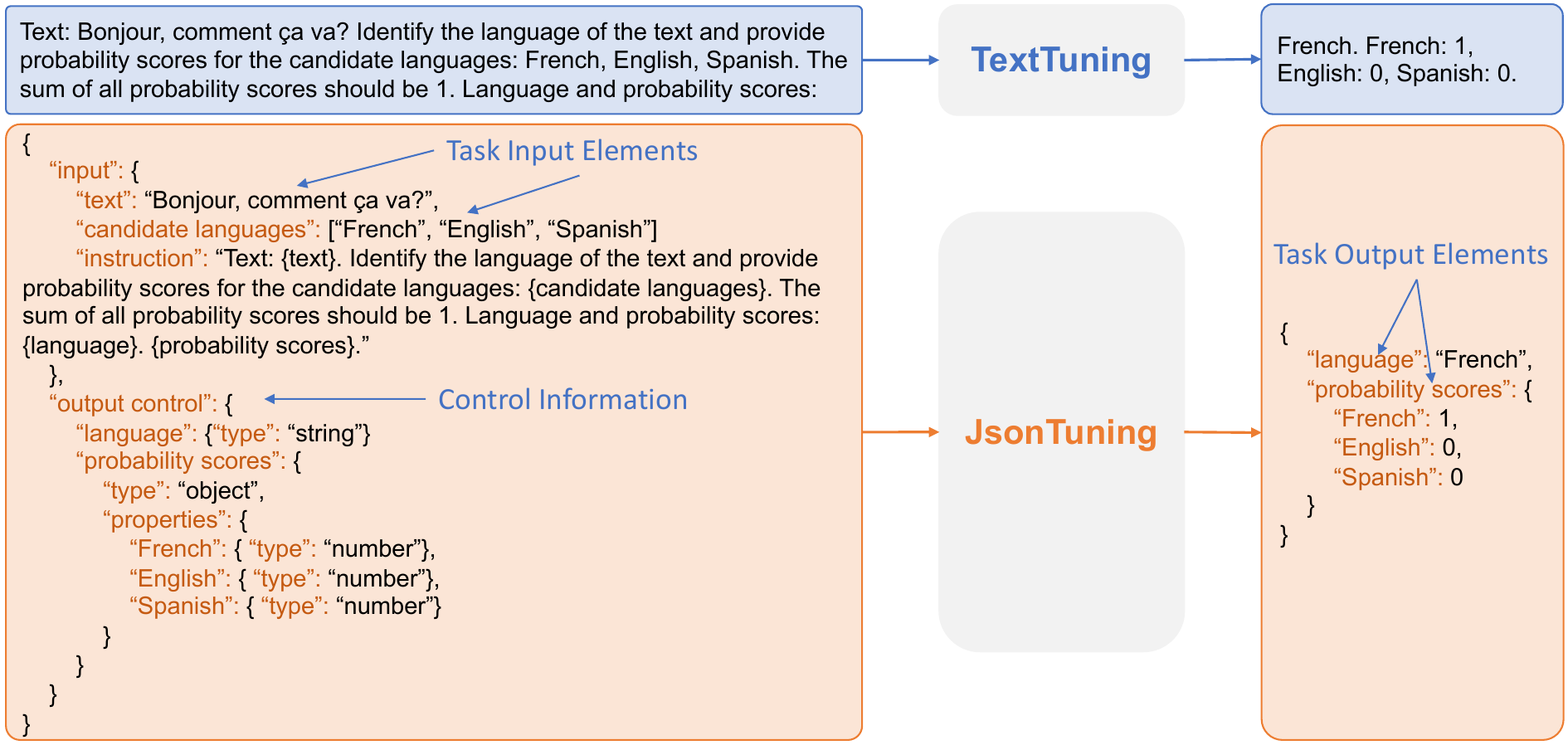}
    \caption{Overview of the typical TextTuning method and our proposed JsonTuning paradigm.}
    \label{fig:jsontuning}
     \vspace{-3mm}
\end{figure*}

Existing instruction tuning methods formulate all tasks as natural language generation \citep{wei2022finetuned, sanh2022multitask, wang-etal-2022-super, chung2022scaling},
a strategy aligned with the typical pretraining of LLMs on language modeling tasks.
However, natural language instructions can be ambiguous, leading to suboptimal understanding or unintended outputs from the model, especially for complex tasks.
Specifically, such text-to-text instruction tuning (TextTuning) methods suffer from the following limitations: (1) \textbf{Generalization}. As presented in Figure \ref{fig:jsontuning}, TextTuning methods mix task elements (e.g., \emph{text} and \emph{candidate languages}) and instructions in natural language texts, which can obscure the structure in tasks. This lack of explicit task structure may introduce ambiguity in essential task elements and their relations, potentially hindering models' generalization abilities. 
 (2) \textbf{Robustness}. Ambiguity in natural language texts can lead to models being sensitive to input variations, resulting in a lack of robust performance. TextTuning methods have been shown sensitive to phrasings of instructions \citep{sanh2022multitask, sun2023evaluating}, variations of labels \citep{ye2023context, wei2023symbol}, and the order of options \citep{pezeshkpour2023large, zheng2024large}. 
 (3) \textbf{Controllability}. It can be difficult to provide a clear description or enforce a specific structure for the desired output due to the ambiguity of natural language \citep{han2023information}, preventing the model from effectively controlling the output.

To address the above limitations, it is crucial to incorporate explicit task structure into the input and output representations during instruction tuning. Structured data representations such as JavaScript Object Notation (JSON) can mitigate misunderstandings and enhance clarity regarding task objectives. In this paper, we introduce JsonTuning, a novel structure-to-structure approach leveraging the versatility and structured nature of JSON for instruction tuning. 
The key idea is to represent the inputs and outputs of all tasks as JSON structures, with the input JSON structure containing task input elements, instructions, and control information, and the output JSON structure comprising task output elements.

We compare JsonTuning with TextTuning in Figure \ref{fig:jsontuning}. JsonTuning addresses the limitations of TextTuning in the following ways: 
(1) \textbf{Generalization}. By explicitly representing the structure in tasks, JsonTuning enhances the model's understanding of essential task elements and their underlying relations and ensures a consistent representation of data across different tasks, leading to improved generalization and adaptability to new tasks.
(2) \textbf{Robustness}. JsonTuning helps minimize ambiguity and manage inconsistencies in the data, facilitating the model to process and generate accurate outputs when faced with input variations, resulting in enhanced robustness. 
(3) \textbf{Controllability}. JsonTuning offers explicit control over the output structure and content, enabling the model to effectively manage output generation. For the language detection task in Figure \ref{fig:jsontuning}, JsonTuning clearly describes the output structure, including the organization and data types of output elements, which is challenging or even impossible to achieve using natural language texts alone.

We conduct a comparative study to demonstrate the advantages of JsonTuning by instruction-tuning various pre-trained language models
and assessing the performance of JsonTuning and TextTuning in terms of generalization, robustness, and controllability across a diverse range of tasks, including representative benchmarks MMLU \citep{MMLU} and BBH \citep{BBH}, tasks with intricate input and output structures, and open-ended instruction-following tasks.
The experimental results reveal the following key findings: (1) JsonTuning consistently outperforms TextTuning in terms of generalization across all language models and tasks, with average performance improving from 26.78 to 30.88. (2) Json-tuned models exhibit significantly greater robustness compared to Text-tuned models with respect to variations in instructions and labels. (3) Json-tuned models demonstrate the ability to generalize to more complex structures when trained on a limited number of simpler structured tasks and generate the desired output in a well-defined structured format.

\section{JsonTuning: Structure-to-Structure Instruction Tuning}

\begin{table*}[t]
\begin{center}
\resizebox{\textwidth}{!}{
\begin{tabular}{p{0.08\linewidth}  p{0.78\linewidth} p{0.22\linewidth}}
 \textbf{Method} & \textbf{Input} & \textbf{Output}  \\ 
\toprule
Prompt & \blue{Answering the following question: \{question\} \{options\}. Answer:} & \orange{\{answer\}} \\
\midrule
Example &  \{ 
        \jsonkey{``question''}: \jsonvalue{``Who is the CEO of Google?''},
        \jsonkey{``options''}: \jsonvalue{``(A) Sundar Pichai (B) Bill Gates (C) Tim Cook (D) Satya Nadella''} \}  & \{ \jsonkey{``answer''}: ``(A)'' \} \\
\midrule
Text & Output Control: The output consists of an answer, which is a string. Answering the following question: Who is the CEO of Google? (A) Sundar Pichai (B) Bill Gates (C) Tim Cook (D) Satya Nadella. Answer: & (A) \\ 
\midrule
Json & \{\jsonkey{``input''}: \{ 
        \jsonkey{``question''}: \jsonvalue{``Who is the CEO of Google?''},
        \jsonkey{``options''}: \jsonvalue{``(A) Sundar Pichai (B) Bill Gates (C) Tim Cook (D) Satya Nadella''},
        \jsonkey{``instruction''}: \jsonvalue{``Answering the following question: \{question\} \{options\}. Answer: \{answer\}''}
    \}, 
    
    \jsonkey{``output control''}: \{
        \jsonkey{``answer''}: \{
            \jsonkey{``type''}: \jsonvalue{``string''}
        \}
    \}
\}  & \{ \jsonkey{``answer''}: ``(A)'' \}
 \\ 
\bottomrule
\end{tabular}
}
\caption{Multiple-choice question answering (MCQA) examples of TextTuning and JsonTuning. The task prompt consists of an input template and an output template, which are highlighted in \blue{blue} and \orange{orange}, respectively.}
\label{tab:method}
\end{center}
\vspace{-4mm}
\end{table*}

\subsection{Unified Structure-to-Structure Formulation}
\label{section:data_representation}

We formulate instruction tuning as a structure-to-structure generation problem, representing task inputs and outputs as JSON structures. Given a task $T$, we denote its input elements as $T_I = (I_1, I_2, \dots, I_n)$ and output elements as $T_O = (O_1, O_2, \dots, O_m)$, where $I_i$ is the $i$th input element, and $O_i$ is the $i$th output element.
Taking the multiple-choice question answering (MCQA) task in Table \ref{tab:method} for illustration, it has two input elements: \textit{question} and \textit{options} and an output element: \textit{answer}. 
With $T_I$, $T_O$, the task prompt $TP$, and control information $C$, we construct the input JSON structure $S_I$ and output JSON structure $S_O$ as follows:

$S_I = \{\text{``input''}: \{I_1: v_1, \dots, I_n: v_n, \text{``instruction''}: TP\}, 
\text{``output control''}: C\}$

$S_O = \{O_1: u_1, \dots, O_m: u_m\}$

\noindent where $v_i$ is the value of $I_i$, and $u_i$ is the value of $O_i$. We identify the following components for effective instruction tuning:
\begin{itemize}[leftmargin=0.4cm]
    \item \textbf{Task Prompt $TP$}. The task prompt $TP$ provides instructions for generating $T_O$ conditioned on $T_I$ and is necessary for instruction tuning. We incorporate a key named \emph{instruction} in $S_I$ to provide such information.

    \item \textbf{Control Information $C$}. The control information $C$ specifies the structured format, explanations, and constraints for the output. We employ JSON Schema to define $C$, resulting in $C$ being a JSON structure as well. Incorporating $C$ into $S_I$ provides the following advantages: (1) \emph{Enhancing controllability}. JSON Schema allows $C$ to precisely define the expected output. As presented in Figure \ref{fig:jsontuning}, the control information for the language detection task indicates that the output consists of two elements. The first element, ``language,'' is defined as a string, while the second element, ``probability scores,'' is defined as an object containing three properties, each being a number. Precisely describing such structure and constraints for the output using natural language texts poses a considerable challenge.
    (2) \emph{Improving generalization to new structures}. Integrating $C$ enables the model to learn the relationship between the constraints in $C$ and the corresponding values in $S_O$, allowing it to generalize to new combinations of basic components. 
    (3) \emph{Increasing training consistency}. Different tasks may require varying output structures, and even a single task may have different output structures. Without $C$, the model might struggle to understand how to map a specific input to the appropriate output structure.
    
\end{itemize}

With $S_I$ and $S_O$, we can employ a language model $M: S_I \rightarrow S_O$ for training and inference.

\subsection{Data Source}

The Flan 2022 collection \citep{chung2022scaling, longpre2023flan} is a comprehensive and widely-used public instruction tuning collection consisting of over 1800 tasks. It integrates resources from Flan 2021 \citep{wei2022finetuned}, P3++ \citep{sanh2022multitask}, Super-Natural Instructions \citep{wang-etal-2022-super}, and additional reasoning, dialogue, and program synthesis datasets. For our primary experiments, we randomly sample a subset from the Flan 2022 collection, maintaining the original collection's data proportion to ensure task diversity. 

Despite the diverse tasks in the Flan 2022 collection, the input and output structures are relatively simple. The outputs for nearly all tasks are purely textual, lacking arrays, objects, or their combinations. Consequently, language models tuned with the Flan 2022 collection may struggle to generalize to diverse and complex structured tasks. To address this limitation, we introduce structured tasks for instruction tuning. Specifically, we employ information extraction (IE) tasks from InstructUIE \citep{wang2023instructuie} as structured tasks for the following reasons: (1) they are well-defined and representative, as numerous structure prediction tasks, such as semantic role labeling and coreference resolution, can be formulated as IE tasks \citep{TANL, wang-etal-2022-deepstruct}; 
(2) they possess complex input and output structures; (3) different IE task datasets have varying schemas, such as different entity categories and relations, thus fostering diversity.
InstructUIE comprises three tasks: named entity recognition (NER), relation extraction (RE), and event extraction (EE). We utilize the NER and RE tasks for training, reserving the EE task for evaluation. Since the output structure of the EE task is more intricate than that of the NER and RE tasks, we can assess the instruction-tuned language models' capability to generalize to more complex structures. To encourage diversity, we uniformly select examples from the training sets of multiple datasets of each task for tuning. Further details regarding the training datasets of IE tasks can be found in Appendix \ref{section:IE_datasets}.

\subsection{Data Representation}

We use the defined data structures $S_I$ and $S_O$ in Section \ref{section:data_representation} to represent all tuning data in JSON structured format with the following data types: \texttt{object}, \texttt{array}, and \texttt{string}. The \texttt{number} and \texttt{boolean} types can be represented as the \texttt{string} type for simplicity. Further details regarding JSON and its utilization are available in Appendix \ref{section:json}.

The structures $S_I$ and $S_O$ can be automatically constructed based on task elements and prompts. Following the approach in \citep{chung2022scaling, sanh2022multitask, wei2022finetuned}, we employ multiple prompts for each task during instruction tuning, and these prompts are evenly distributed across different task examples.
Each prompt $TP$ consists of an input template and an output template. As shown in Table \ref{tab:method}, in the case of an MCQA prompt, the input template could be \textit{``Answer the following question: \{question\} \{options\}. Answer:''}, and the output template could be \textit{``\{answer\}''}. The prompt clearly indicates the essential task elements, namely \textit{question}, \textit{options}, and \textit{answer}, as well as their relations.
The tasks in the Flan 2022 collection already have multiple prompts. We manually construct 10 prompts each for NER and RE tasks for training, which can be found in Appendix \ref{section:IE_prompts}.
For the control information $C$, all output elements of tasks in the Flan 2022 collection are of the \texttt{string} type, and we manually define $C$ for IE tasks, which can be found in Appendix \ref{section:evaluation}. 

\subsection{Training Data Construction}
The datasets in the Flan 2022 collection and InstructUIE are formatted in JSON. As presented in Table \ref{tab:method}, placeholders within the prompt are substituted with specific task elements from each example to create training instances for TextTuning. By comparison, for JsonTuning, task elements are directly incorporated into the input and output JSON structures, with the ``instruction'' key serving to denote the prompt. To integrate control information within TextTuning, we can employ natural language texts to describe the corresponding JSON schema.

\section{Experiments}
\label{sec:experiments}

\begin{table*}[t]
\begin{center}
\setlength\tabcolsep{3.5pt}
\begin{tabular}{ccccccccl}
\textbf{Model}  & \textbf{Method} & \textbf{MMLU} & \textbf{BBH} & \textbf{NER} & \textbf{RE} & \textbf{EE} &  \textbf{NL2SQL} & \textbf{Average} \\ 
\toprule
\multirow{2}{*}{Falcon-7B}  
& Text & 24.64 & 20.64 & 20.72 & 6.05 & 0.33 / 0.00 & 2.00 & 12.37 \\ 
& Json & 34.13 & 32.61 & 29.02 & 8.36 & 0.31 / 0.28 & 1.40 & \textbf{17.64} \\
\midrule
\multirow{2}{*}{Mistral-7B}  
& Text & 51.42 & 40.09 & 40.18 & 22.01 & 2.37 / 0.00 & 30.80 & 30.95 \\ 
& Json & 51.79 & 41.79 & 53.02 & 25.09 & 7.26 / 14.63 & 31.80 & \textbf{35.74} \\
\midrule
\multirow{2}{*}{LLaMA-7B}   
& Text & 43.11 & 32.48 & 37.61 & 14.33 & 1.35 / 0.00 & 8.60  & 22.80 \\ 
& Json & 44.69 & 37.08 & 43.47 & 15.28 & 3.49 / 7.33 & 16.40 & \textbf{27.06} \\
\midrule
\multirow{2}{*}{LLaMA-13B}  
& Text & 49.49 & 39.07 & 38.15 & 21.40 & 1.70 / 0.00 & 17.80 & 27.79 \\ 
& Json & 48.98 & 40.47 & 45.31 & 22.97 & 4.20 / 10.73 & 21.40 & \textbf{31.10} \\
\midrule
\multirow{2}{*}{LLaMA2-7B}  
& Text & 46.36 & 37.89 & 41.66 & 20.74 & 0.55 / 0.00 & 10.80 & 26.29 \\
& Json & 47.95 & 39.23 & 45.25 & 23.98 & 4.23 / 10.81 & 11.20 & \textbf{29.19} \\
\midrule
\multirow{2}{*}{LLaMA2-13B}  
& Text & 52.30 & 41.91 & 40.95 & 22.51 & 1.48 / 0.00 & 23.20 & 30.27 \\
& Json & 51.88 & 42.85 & 47.71 & 23.18 & 6.65 / 11.00 & 26.40 & \textbf{33.47} \\
\midrule
\multirow{2}{*}{LLaMA3-8B}  
& Text & 58.22 & 44.49 & 43.31 & 22.34 & 1.34 / 0.00 & 53.00 & 37.01 \\ 
& Json & 59.24 & 46.77 & 53.15 & 27.33 & 7.67 / 16.44 & 53.20 & \textbf{41.96} \\
\midrule
\multirow{2}{*}{Average}
& Text & 46.51 & 36.65 & 37.51 & 18.48 & 1.30 / 0.00 & 20.89 & 26.78 \\ 
& Json & \textbf{48.38} & \textbf{40.11} & \textbf{45.28} & \textbf{20.89} & \textbf{4.83} / \textbf{10.17} & \textbf{23.11} & \textbf{30.88} \\
\bottomrule
\end{tabular}
\end{center}
\caption{Generalization results on diverse benchmarks and tasks. For simplicity, we refer to JsonTuning and TextTuning as ``Json'' and ``Text'', respectively.}
\label{tab:generalization}
 \vspace{-4mm}
\end{table*}

\subsection{Experimental Setup}

\paragraph{Pre-trained Language Models}

We adopt seven prevalent pre-trained language models, namely Falcon-7B \citep{Falcon}, Mistral 7B \cite{mistral}, LLaMA-7B, LLaMA-13B \citep{LLaMA}, LLaMA2-7B, LLaMA2-13B \citep{LLaMA2}, and LLaMA3-8B \cite{llama3}, for our experiments. These models are trained on trillions of tokens and are among the most widely used open-source language models.

\paragraph{Evaluation Tasks and Datasets}
We focus on performance on unseen tasks and datasets.
 We evaluate models on popular aggregated benchmarks: MMLU \citep{MMLU} consisting of 57 tasks of exam questions and BBH \citep{BBH} including 23 challenging tasks from BIG-Bench \citep{srivastava2023beyond} following \citet{chung2022scaling}. In addition, we adopt tasks with complex input and output structures for evaluation. Specifically, we use the NER, RE, and EE tasks from InstructUIE \citep{wang2023instructuie} and the NL2SQL task which requires the conversion of natural language queries into SQL using a provided structured database schema consisting of table names and column names.
 For NER and RE, we use datasets unseen during training. Specifically, we use 5 datasets, namely, AI, literature, music, politics, and science, from CrossNER \citep{CrossNER} for the NER task and 2 datasets, namely CoNLL2004 \citep{conll04} and FewRel \citep{FewRel}, for the RE task. For the unseen EE task, we use ACE2005 \citep{ace2005-annotation}, CASIE \citep{Satyapanich_Ferraro_Finin_2020}, and PHEE \citep{PHEE} datasets for evaluation. We use the Spider \citep{yu-etal-2018-spider} dataset for NL2SQL. Apart from datasets in MMLU and BBH, we randomly select up to 500 examples for each dataset from its test set for evaluation so that a single dataset will not dominate the results of its task and the evaluation cost is acceptable. The details of evaluation datasets and prompts are in Appendix \ref{section:evaluation}. 

\paragraph{Evaluation Metrics}
We use accuracy for MMLU and BBH following \citet{chung2022scaling}, entity F1 for the NER task, relation boundary F1 for the RE task, event trigger F1 and argument F1 for the EE task following \citet{wang2023instructuie}, and execution accuracy for NL2SQL following \citet{yu-etal-2018-spider}. 

\paragraph{Implementation Details}
We employ the parameter-efficient method LoRA (Low Rank Adaptation) \citep{hu2022lora} for fine-tuning. We use 50K examples from the Flan collection 2022 and 10K examples from structured tasks in InstructUIE, with an equal division between the NER and RE tasks, and train the learnable parameters for 3 epochs with a batch size of 64. For model optimization, we use the AdamW \citep{AdamW} optimizer with linear learning rate decay, and the peak learning rate is set to 1e-3. We set the maximum length as 2048 for training and evaluation. For evaluation, we use greedy decoding in all scenarios. We conduct training and evaluation of diverse language models employing both JsonTuning and TextTuning methods on identical datasets. This approach facilitates a fair and direct comparison between JsonTuning and TextTuning. 

\subsection{Generalization Results}

\begin{figure*}[ht]
    \centering
    \includegraphics[width=0.85\linewidth]{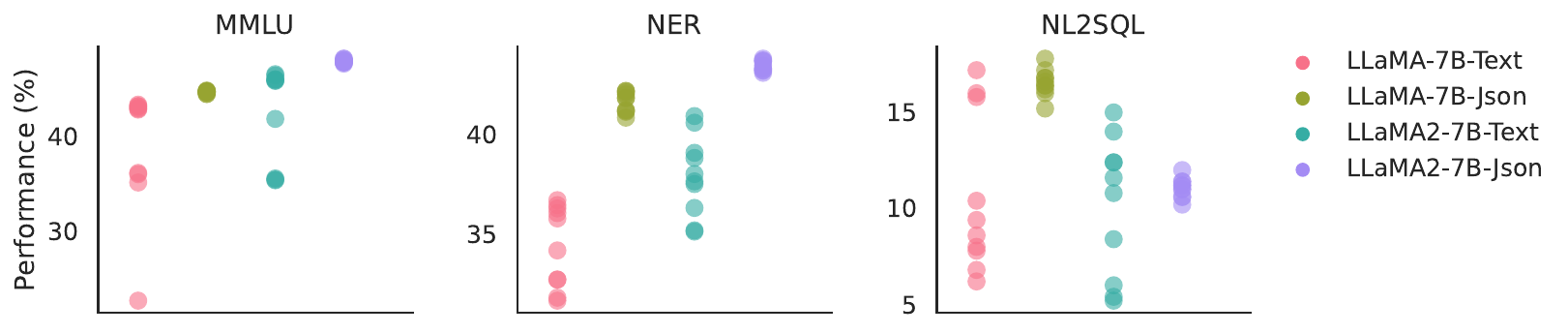}
    \caption{Performance of JsonTuning and TextTuing models with different prompts. Each point in the figure represents the performance associated with the application of a particular prompt.}
    \label{fig:prompt}
     \vspace{-3mm}
\end{figure*}

Table \ref{tab:generalization} presents the zero-shot generalization results of JsonTuning and TextTuning using five language models.  We have the following observations:

\begin{itemize}[leftmargin=0.4cm]
    \item \textit{JsonTuning surpasses TextTuning in the majority of tasks and models.} This is evident from the higher average scores for JsonTuning across all models and tasks, where JsonTuning achieves an overall average score of 30.88 compared to TextTuning's 26.78. This suggests that JsonTuning is a more effective method for instruction tuning.
    \item \textit{JsonTuning significantly improves the model's ability to tackle complex structured tasks.} JsonTuning consistently outperforms TextTuning on tasks with complex structures, such as NER, EE, and NL2SQL. Json-tuned models can adapt to intricate EE structures, even when only trained on simpler NER and RE structures. In contrast, Text-tuned models rarely generate valid EE structures. These observations demonstrate the superior controllability and generalization ability of JsonTuning.
    \item \textit{JsonTuning allows models to better leverage its abilities and knowledge when responding to human instructions, particularly for models with limited capabilities.} For example, Falcon-7B with JsonTuning exhibits a substantial improvement over TextTuning on tasks such as MMLU and BBH, highlighting the importance of an appropriate instruction-tuning method for unlocking the model's potential.
    
\end{itemize}

\subsection{Robustness Results}

The robustness of instruction-tuned language models is of paramount importance for their successful deployment across a diverse range of tasks. In this section, we assess the model's resilience against varying prompts and unseen labels, which have been identified as challenging aspects for instruction-tuned models in prior research \citep{sanh2022multitask, sun2023evaluating, ye2023context}. 

\begin{figure}[ht]
    \centering
    \includegraphics[width=0.96\linewidth]{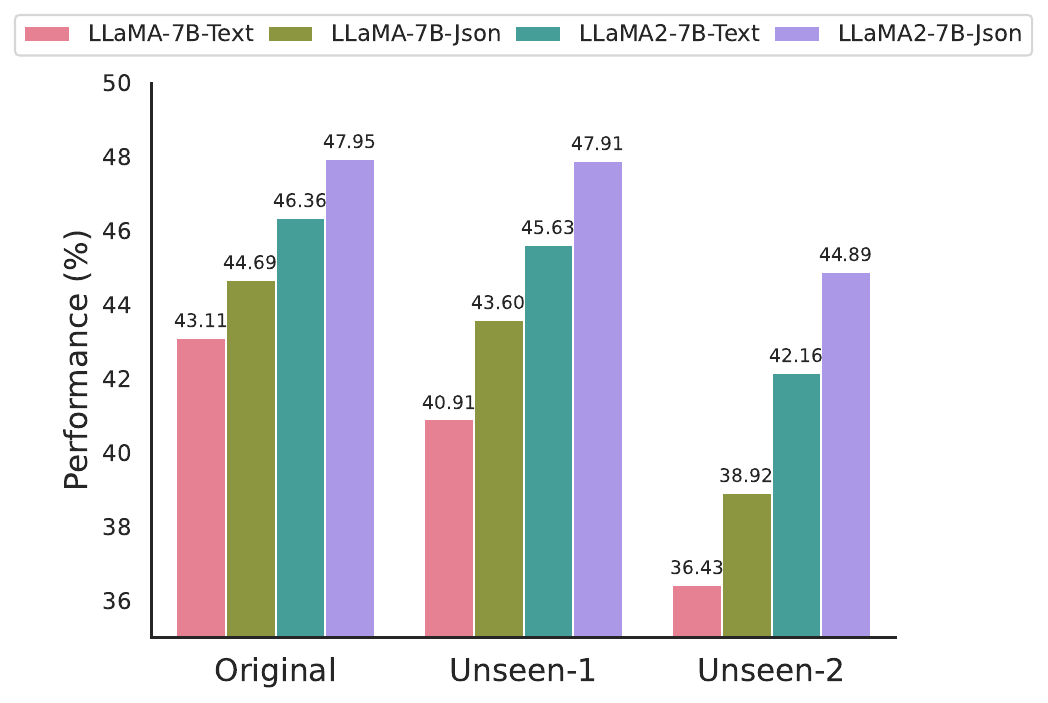}
    \caption{Performance of JsonTuning and TextTuing models with different label spaces on MMLU.}
    \label{fig:label}
     \vspace{-3mm}
\end{figure}

To evaluate prompt robustness, we employ 10 distinct prompts for the MMLU benchmark, the NER task, and the NL2SQL task. Detailed information is in Appendix \ref{section:evaluation}. Figure \ref{fig:prompt} illustrates the performance of models trained with JsonTuning and TextTuning on these tasks when subjected to different prompts. Our findings reveal that Json-tuned models exhibit greater robustness compared to Text-tuned models, as evidenced by higher mean performance and reduced variance. 
In terms of label robustness, we substitute the MMLU label space with previously unseen label spaces. The original label space for MMLU is \texttt{\{(A), (B), (C), (D)\}}, with these option letters frequently encountered in the training dataset. We replace this label space with two alternatives: \texttt{\{(W), (X), (Y), (Z)\}} and \texttt{\{(\$), (€), (£), (¥)\}}, denoted by Unseen-1 and Unseen-2, respectively. These label spaces were not present during instruction tuning.
As shown in Figure \ref{fig:label}, Json-tuned models consistently outperform Text-tuned models in all scenarios. 
These improvements can be attributed to JsonTuning's ability to effectively differentiate between instructions and task elements, thereby minimizing ambiguity and enhancing robustness.


\subsection{Controllability Results}
\paragraph{Case studies on controllability}

\begin{table*}[t]
\begin{center}
\resizebox{\textwidth}{!}{
\begin{tabular}{p{0.08\linewidth}  p{0.77\linewidth} p{0.26\linewidth}}
 \textbf{Method} & \textbf{Input} & \textbf{Prediction}  \\ 
\toprule
\multicolumn{3}{c}{\textbf{Task: Language detection with probability scores}}  \\
\midrule
Text &  Text: Bonjour, comment ça va? Identify the language of the text and provide probability scores for the candidate languages: French, English, Spanish. The sum of all probability scores should be 1. Language and probability scores: & French .7
 \\ 
\midrule
Json &  \{
    \jsonkey{``input''}: \{
        \jsonkey{``text''}: \jsonvalue{``Bonjour, comment ça va?''},
        \jsonkey{``candidate languages''}: \jsonvalue{[``French'', ``English'', ``Spanish'']},
        \jsonkey{``instruction''}: \jsonvalue{``Text: \{text\}. Identify the language of the text and provide probability scores for the candidate languages: \{candidate languages\}. The sum of all probability scores should be 1. Language and probability scores: \{language\} \{probability scores\}''}
    \},

    \jsonkey{``output control''}: \{
        \jsonkey{``language''}: \{\jsonkey{``type''}: \jsonvalue{``string''}\},
        \jsonkey{``probability scores''}: \{
            \jsonkey{``type''}: \jsonvalue{``object''},
            \jsonkey{``properties''}: \{
                \jsonkey{``French''}: \{\jsonkey{``type''}: \jsonvalue{``string''} \},
                \jsonkey{``English''}: \{\jsonkey{``type''}: \jsonvalue{``string''} \},
                \jsonkey{``Spanish''}: \{\jsonkey{``type''}: \jsonvalue{``string''} \}
            \}
        \}
    \}
\} & \{\jsonkey{``language''}: \jsonvalue{``French''}, \jsonkey{``probability scores''}: \{ \jsonkey{``French''}: \jsonvalue{0.98}, \newline\jsonkey{``English''}: \jsonvalue{0.01}, \newline \jsonkey{``Spanish''}: \jsonvalue{0.01}\} \} \\
\midrule
\multicolumn{3}{c}{\textbf{Task: Intent detection and slot filling}}  \\
\midrule
Text & Text: Set an alarm for 7 AM tomorrow. Detect the intent of the text and extract time and date slots from the text: & Set an alarm for 7 AM tomorrow
 \\ 
\midrule
Json &  \{
    \jsonkey{``input''}: \{
        \jsonkey{``text''}: \jsonvalue{``Set an alarm for 7 AM tomorrow.''},
        \jsonkey{``instruction''}: \jsonvalue{``Text: \{text\}. Detect the intent or purpose of the text and extract time and date slots from the text:''}
    \},
    
    \jsonkey{``output control''}: \{
        \jsonkey{``intent''}: \{\jsonkey{``type''}: \jsonvalue{``string''}\},
        \jsonkey{``slots''}: \{
            \jsonkey{``type''}: \jsonvalue{``object''},
            \jsonkey{``properties''}: \{
                \jsonkey{``time''}: \{\jsonkey{``type''}: \jsonvalue{``string''}\},
                \jsonkey{``date''}: \{\jsonkey{``type''}: \jsonvalue{``string''}\}
            \}
        \}
    \}
\} & \{\jsonkey{``intent''}: \jsonvalue{``setAlarm''}, \jsonkey{``slots''}: \{\jsonkey{``time''}: \jsonvalue{``7:00''}, \jsonkey{``date''}: \jsonvalue{``tomorrow''}\} \} \\
\bottomrule
\end{tabular}
}
\end{center}
\caption{Case studies focusing on controllability. Each example displays its input along with the model's prediction.}
\label{tab:control}
\vspace{-4mm}
\end{table*}

In previous sections, we have demonstrated that Json-tuned models possess the capacity to control the output and generalize across complex structures. In this section, we present case studies to qualitatively illustrate the controllability of Json-tuned models. For this purpose, we utilize LLaMA2-13B trained with both JsonTuning and TextTuning approaches.
As evidenced by Table \ref{tab:control}, JsonTuning effectively enables the model to identify the desired output, generating results in a well-structured format. In contrast, the Text-tuned model fails to adequately adhere to the provided instructions. 
For example, in the language detection task, the Text-tuned model struggles to provide clear probability scores. The output, such as ``.7'', is ambiguous and difficult to interpret. By comparison, the Json-tuned model successfully follows the instruction, delivering scores that meet the specified requirements. Additional case studies are in Appendix \ref{section:control}.

\paragraph{Quantitative assessment of controllability}
To rigorously evaluate the controllability of Json-tuned models, we analyze the proportion of invalid JSON structures generated by Json-tuned LLaMA-7B and LLaMA2-7B across a diverse set of evaluation tasks. Additionally, we assess whether valid JSON outputs comply with the specified control information. Our findings reveal that all valid JSON outputs consistently adhere to the provided control information across the evaluated tasks. Meanwhile, as shown in Table \ref{tab:invalid}, the occurrence of invalid JSON structures is infrequent for both models. These observations collectively highlight the reliability and controllability of our JsonTuning approach in handling JSON-formatted data.

\begin{table*}[ht]
\begin{center}
\begin{tabular}{ccccccccc}
 \textbf{Model} & \textbf{MMLU} & \textbf{BBH} & \textbf{NER} & \textbf{RE} & \textbf{EE} & \textbf{NL2SQL}  \\ 
\toprule
LLaMA-7B & 0.00\% & 0.74\% & 0.00\% & 0.10\% & 0.23\% & 0.20\% \\
LLaMA2-7B & 0.00\% & 0.48\% & 0.30\% & 0.00\% & 0.85\% & 2.00\% \\
\bottomrule
\end{tabular}
\caption{Proportions of invalid JSON structures on different evaluation tasks.}
\label{tab:invalid}
\vspace{-2mm}
\end{center}
\end{table*}

\section{Analysis}
\label{sec:analysis}
\paragraph{Does JsonTuning bring benefits in open-ended instruction-following scenarios?}

\begin{figure}[t]
    \centering
    \includegraphics[width=0.9\linewidth]{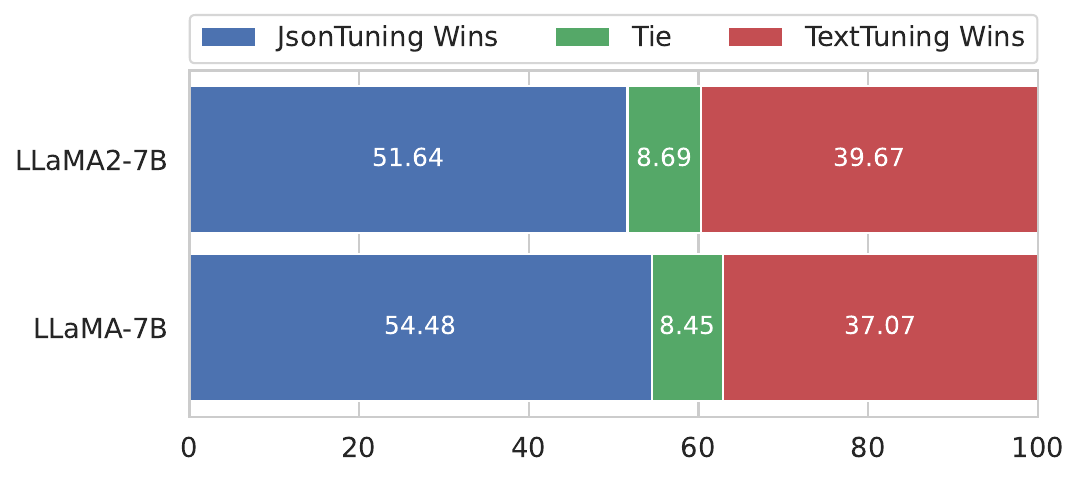}
    \caption{Preference evaluation on AlpacaEval using GPT-4 as the evaluator.}
    \label{fig:alpacaeval}
    \vspace{-3mm}
\end{figure}

While performance on the aforementioned benchmarks effectively quantifies the models' capabilities in specific skills, these metrics may not accurately reflect the models' proficiency in handling open-ended instructions.
To investigate the potential advantages of JsonTuning in such scenarios, we continue our training of Json-tuned and text-tuned models on the Alpaca dataset \cite{alpaca} and assess their performance on AlpacaEval \cite{alpaca_eval}, with LLaMA-7B and LLaMA2-7B models being used for our experiments. The evaluation prompt and examples from AlpacaEval are provided in Figure \ref{fig:alpacaeval_prompt_examples}.
As illustrated in Figure \ref{fig:alpacaeval}, JsonTuning also demonstrates significant advantages over TextTuning in open-ended instruction-following scenarios. This benefit likely stems from JsonTuning's enforcement of a consistent and standardized data representation. This structured approach mitigates the risk of misinterpretations that often occur due to textual variations in traditional text-based tuning.

\paragraph{What is the importance of explicit representation of task structures?}
To underscore the significance of explicitly representing task structures in the input, we conduct an ablation study comparing two methods: Text2Json and Json2Json (i.e., our proposed JsonTuning method). In the Text2Json approach, we use a text-based input, structured control information, and a structured output. The key distinction between Text2Json and Json2Json is that Text2Json relies on a text input and does not explicitly represent task structures.
The results, as shown in Table \ref{tab:text2json}, reveal a clear performance advantage for our Json2Json approach over Text2Json, indicating that explicitly representing task structures in the input is crucial, and structured output alone is insufficient.

\begin{table*}[t]
\begin{center}
\setlength\tabcolsep{3.5pt}
\begin{tabular}{ccccccccc}
 \textbf{Model} &  \textbf{Method} & \textbf{MMLU} & \textbf{BBH} & \textbf{NER} & \textbf{RE} & \textbf{EE} & \textbf{NL2SQL} & \textbf{Average} \\ 
\toprule
\multirow{2}{*}{LLaMA-7B} 
& Text2Json & 41.92 & 36.15 & 36.38 & 14.44 & 2.80 / 5.35 & 13.00 & 24.33 \\ 
& Json2Json & 44.69 & 37.08 & 43.47 & 15.28 & 3.49 / 7.33 & 16.40 & \bf{27.06} \\
\midrule
\multirow{2}{*}{LLaMA2-7B} 
& Text2Json & 47.29 & 38.14 & 41.49 & 16.99 & 4.07 / 10.85 & 11.40 & 27.13 \\
& Json2Json & 47.95 & 39.23 & 45.25 & 23.98 & 4.23 / 10.81 & 11.20 & \textbf{29.19} \\
\bottomrule
\end{tabular}
\caption{Performance Comparison Between Text2Json and Json2Json. Text2Json uses a text input along with structured control information and output, without explicitly encoding task structures in the input. Json2Json (i.e., our proposed JsonTuning method), on the other hand, incorporates explicit task structures in the input.}
\label{tab:text2json}
\end{center}
\end{table*}

\section{Related Work}
The emergence of large language models (LLMs) has had a transformative effect on the AI field. These advancements have led to a surge in the release of both closed-source LLMs \citep{Chatgpt, GPT-4, Claude3} and open-source LLMs \citep{zhang2022opt, Falcon, LLaMA, LLaMA2, bai2023qwen, mistral}, fostering innovation and collaboration within the research community.
Instruction tuning \citep{wei2022finetuned, mishra-etal-2022-cross,  sanh2022multitask, opt-iml, chung2022scaling, wang-etal-2023-self-instruct, alpaca} has emerged as a promising research direction, leveraging the capabilities of LLMs to enhance their responsiveness to human instructions. Collections such as Super-NaturalInstructions \citep{wang-etal-2022-super}, the Flan 2022 collection \citep{chung2022scaling}, and open-ended instruction-following datasets \cite{alpaca, kopf2023openassistant, dolly, peng2023instruction, ultrachat, xu2023wizardlm} have accelerated the development of instruction-tuned models. 

To advance instruction tuning, researchers have explored learning from human feedback \citep{summarization, InstructGPT, bai2022training, scheurer2023training}, automatic data generation \citep{wang-etal-2023-self-instruct, peng2023instruction, xu2023wizardlm, yin-etal-2023-dynosaur}, and data selection \citep{zhou2023lima, cao2023instruction, lu2023instag, liu2023makes}. While the learning algorithm and tuning data have received considerable attention from researchers, the significance of data representation has often been overlooked. Our JsonTuning approach offers an alternative perspective on data representation to enhance instruction tuning in terms of generalization, robustness, and controllability. 

Our approach is significantly distinct from constrained decoding methods, such as those employed by commercial LLM APIs (e.g., OpenAI's JSON mode) or systems relying on human-designed grammars \citep{deutsch-etal-2019-general,shin-etal-2021-constrained,geng-etal-2023-grammar}. These methods, which range from token-level constraints to high-level format specifications, are utilized to ensure models conform to prescribed syntactic and semantic requirements. While JsonTuning shares a similar objective of controlling outputs, its primary focus lies in leveraging task structures throughout the instruction-tuning process. This approach aims to enhance generalization, robustness, and controllability, rather than merely enforcing adherence to a predetermined output format. Furthermore, unlike constrained decoding methods that typically involve modifying the decoding process—adding complexity and potentially increasing latency, our approach is simpler and avoids the need for such modifications.

\section{Conclusion}
This paper introduces JsonTuning, a novel approach designed to overcome the limitations of conventional text-to-text instruction tuning methods for language models. By utilizing the structured data format for explicit task representation, JsonTuning significantly improves the model's generalization, robustness, and controllability. Our experimental results and case studies highlight the benefits of JsonTuning in generalizing to unseen tasks and datasets, maintaining robustness against varying prompts and label spaces, and demonstrating controllability in diverse scenarios.

\section*{Limitations}

Owing to resource constraints, our study focuses on training and evaluating ``mid-sized'' models with less than 20 billion parameters. Investigating the application of JsonTuning on substantially larger language models would be a valuable extension.

JsonTuning is particularly suitable for scenarios requiring high robustness, controllability, or complex structure prediction. We discuss the real-world applications of JsonTuning in Appendix \ref{sec:additional_analysis}, highlighting its benefits derived from robustness, controllability, and explicit task structure representation. Our goal is not to replace TextTuning, as Text-tuned models offer a more natural usage without the need for structured object construction. Instead, our objectives are two-fold: (1) To systematically compare JsonTuning and TextTuning, revealing the inherent limitations of TextTuning in terms of generalization, robustness, and controllability. This comparison provides valuable insights for future research in these areas. (2) To provide a structured data representation and interface across all tasks, offering significant advantages when task structures are utilized.

\section*{Acknowledgements}
This research is supported by the Ministry of Education, Singapore, under its Academic Research Fund (AcRF) Tier 1 grant (SAP 2025\_001).


\appendix

\clearpage
\newpage

\newtcolorbox{mybox}{
colframe = black!75!black,
width = \linewidth,
}

\section{Additional Analysis}
\label{sec:additional_analysis}

\begin{table*}[t]
\begin{center}
\resizebox{\textwidth}{!}{
\setlength\tabcolsep{2pt}
\begin{tabular}{cccccccccc}
 \textbf{Model} &  \textbf{Method} & \textbf{MMLU} & \textbf{BBH} & \textbf{NER}$\dag$ & \textbf{RE}$\dag$ & \textbf{EE} & \textbf{NL2SQL} & \textbf{Average} & \textbf{MMLU-Robustness} \\ 
\toprule
\multirow{4}{*}{LLaMA-7B} 
& Text w/ control info & 42.01 & 34.22 & 35.02 & 14.54 & 0.17 / 0.00	& 11.40 & 22.88 & 39.31 $\pm$ 3.91  \\ 
& Json w/ control info & 44.69 & 37.08 & 43.47 & 15.28 & 3.49 / 7.33 & 16.40 & \bf{27.06} & \bf{44.61} $\pm$ 0.11 \\
\cmidrule{2-10}
& Text w/o control info & 43.11 & 32.48 & 37.61 & 14.33 & 1.35 / 0.00 & 8.60 & 22.80 & 38.82 $\pm$ 6.28 \\
& Json w/o control info & 43.94 & 35.42 & 48.08 & 16.91 & 0.59 / 0.26 & 12.40 & \bf{26.20} & \bf{43.47 $\pm$ 0.26} \\
\midrule
\multirow{4}{*}{LLaMA2-7B} 
& Text w/ control info & 47.94 & 38.44 & 46.08 & 17.88 & 0.95 / 0.00 & 9.20 & 26.70 & 45.05 $\pm$ 3.67 \\
& Json w/ control info & 47.95 & 39.23 & 45.25 & 23.98 & 4.23 / 10.81 & 11.20 & \textbf{29.19} & \bf{47.86 $\pm$ 0.17} \\
\cmidrule{2-10}
& Text w/o control info & 46.36 & 37.89 & 41.66 & 20.74 & 0.55 / 0.00 & 10.80 & 26.29 & 42.46 $\pm$ 4.76 \\
& Json w/o control info & 47.81 & 38.71 & 44.42 & 22.38 & 3.53 / 6.39 & 12.80 & \bf{28.51} & \bf{45.76 $\pm$ 0.19} \\
\bottomrule
\end{tabular}
}
\caption{Ablation results for LLaMA-7B and LLaMA2-7B concerning control information. Tasks marked with $\dag$ are seen during training and evaluated with unseen datasets.}
\label{tab:ablation}
\end{center}
\end{table*}

\paragraph{How does control information impact JsonTuning and TextTuning?}
To answer this question, we conduct experiments using LLaMA-7B and LLaMA2-7B with both TextTuning and JsonTuning, with and without these elements. 
From Table \ref{tab:ablation}, we can make the following observations: (1) JsonTuning consistently outperforms TextTuning across all tested scenarios, whether or not control information is incorporated. This finding underscores the inherent benefits of JsonTuning, which leverages a structure-to-structure learning paradigm to effectively utilize task structures and minimize ambiguity.
(2) Generally, the removal of control information may not negatively impact the model's performance on seen tasks such as NER and RE; in fact, the performance on these tasks may even improve. However, doing so does hinder performance on unseen tasks, resulting in a lower average performance, indicating that they aid the model in generalizing to unseen tasks rather than overfitting to seen tasks. Furthermore, its elimination can compromise the model's robustness, as evidenced by a lower average performance and increased variation on MMLU. 
These observations underscore the crucial role of control information in enhancing generalization and robustness.

\begin{figure*}[t]
    \centering
    \includegraphics[width=\linewidth]{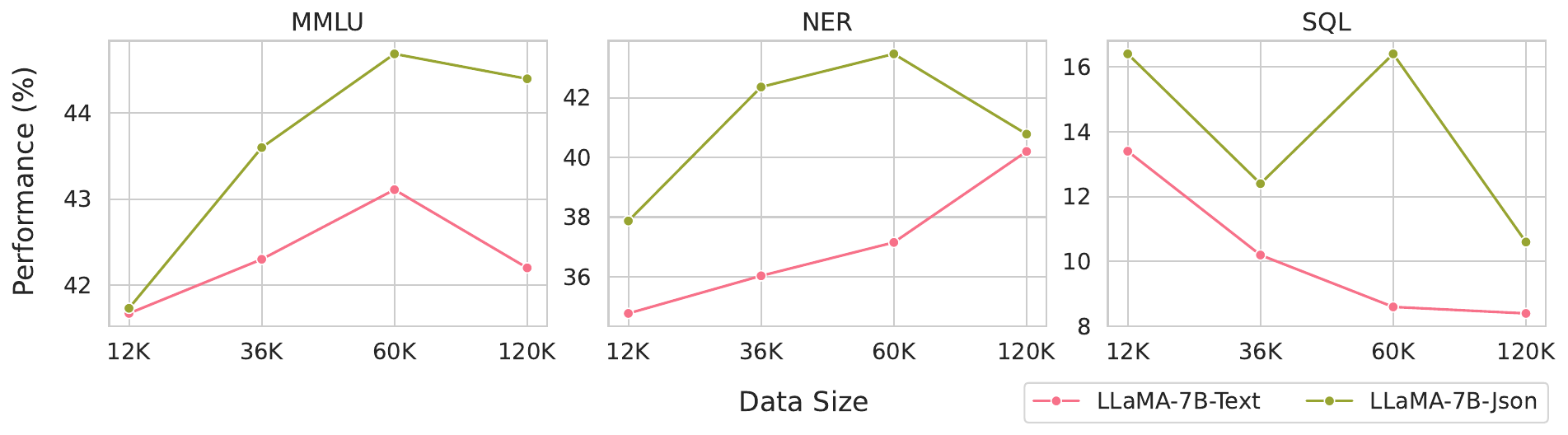}
    \caption{Performance of LLaMA-7B trained using JsonTuning and TextTuning across varying data sizes.}
    \label{fig:data_size}
     \vspace{-2mm}
\end{figure*}

\paragraph{What are the effects of different data sizes on generalization?}
In the primary experiments, we utilize a total of 60K data points, comprising 50K from the Flan 2022 collection and 10K from structured tasks for tuning. In this analysis, we alter the data size while maintaining their relative ratio to examine the effects of different data sizes on generalization. Specifically, we train LLaMA-7B with four different data sizes: 12K, 36K, 60K, and 120K, and evaluate the models on the MMLU benchmark, the NER task, and the NL2SQL task.
Figure \ref{fig:data_size} reveals the following observations:
(1) LLaMA-7B-Json consistently outperforms LLaMA-7B-Text across all tasks and data sizes, indicating the superior generalization capabilities of the JsonTuning model.
(2) Increasing the data size for instruction tuning does not necessarily result in performance improvement, suggesting that enlarging the data size may not be an effective approach to enhance the model's generalization abilities.

\begin{figure*}[ht]
    \centering
    \includegraphics[width=\linewidth]{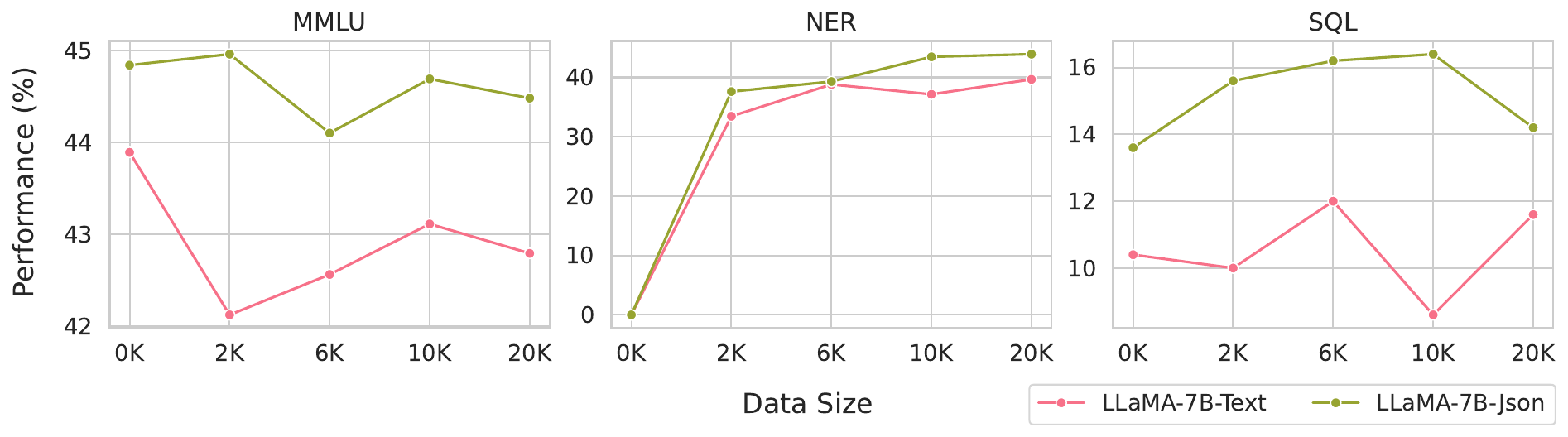}
    \caption{Performance of LLaMA-7B trained using JsonTuning and TextTuning with different numbers of examples of structured tasks.}
    \label{fig:structured_tasks}
\end{figure*}

\paragraph{Are structured tasks essential for instruction tuning?}
To investigate this, we keep the number of examples from the Flan 2022 collection constant and vary the number of examples from structured tasks. Specifically, we use 50K data points from the Flan 2022 collection and 0K, 2K, 6K, 10K, and 20K data points from IE tasks to train LLaMA-7B for the experiments. 
Figure \ref{fig:structured_tasks} reveals the following insights: 
(1) Incorporating structured tasks for training may not enhance the model's generalization ability on tasks without complex structures. Introducing structured tasks for tuning does not improve the model's performance on MMLU, a benchmark without intricate input and output structures.
(2) Structured tasks significantly impact the model's generalization performance on tasks with complex output structures. Without structured tasks for training, the model's performance on the NER task is 0 for both JsonTuning and TextTuning. However, the performance significantly improves when introducing only 2K data points from structured tasks for training. This highlights the importance of structured tasks for instruction-tuned models to generalize to tasks with complex output structures.
(3) Structured tasks have a milder impact on the model's generalization performance on tasks with complex input structures. Introducing an appropriate number of structured tasks can enhance the model's performance on the NL2SQL task, which requires processing a structured database schema. This suggests that training the model with structured tasks aids in processing and understanding complex structures.
In summary, the decision to use structured tasks for instruction tuning depends on the application scenarios. However, regardless of the scenario, JsonTuning consistently appears to be a superior method for instruction tuning compared to TextTuning.

\paragraph{Discussion}
Firstly, our primary claim is that integrating task structures within instruction tuning can enhance its generalization, robustness, and controllability. This claim is not restricted to any specific implementation. We choose JSON due to its robust support for a variety of data types, its straightforward and uniform syntax, and its widespread native support across numerous programming languages, particularly Python. However, other formats like YAML or XML are also viable options, provided that the task structures are explicitly represented.

Secondly, our goal is to provide a structured interface for LLMs. When provided solely with texts, the Json-tuned model may struggle to generate coherent outputs. However, if the textual input is encapsulated within a JSON structure, the model can effectively process the query. JsonTuning modifies the interface without compromising the model's inherent capabilities. While the textual interface may appear more intuitive, it may not be the most effective method for human-model interaction. Introducing a structured interface can provide significant flexibility in various deployment scenarios.
JsonTuning offers several advantages for applications: (1) \emph{Performance}: In complex scenarios, articulating intricate tasks using natural language can be challenging. Conversely, a JSON-structured interface enables users to organize information in a clear and structured manner, significantly enhancing the model's understanding of the task and, consequently, its performance.
(2) \emph{Robustness}: The method's resilience to varying prompts significantly reduces the need for users to test multiple prompts for optimal performance. Furthermore, as backend models are updated or changed, the optimal prompt for natural language texts may vary, necessitating adjustments for each model version. In contrast, Json-tuned models may not require prompt modifications even when the backbone model is updated, considerably reducing user effort. (3) \emph{Controllability}: The output control component allows users to accurately specify and easily parse the output, which can be challenging when using natural language texts. Users only need to comprehend basic types such as string, object, and array in JSON Schema.
To provide both textual and structured interfaces, we may conduct TextTuning and JsonTuning concurrently.

Finally, our methodology is designed to be flexible and adaptable to a variety of structures. The design of the JSON structure can adhere to the following key principles: (1) \emph{Task Identification}: Clearly define the essential task elements, their interrelations, and the anticipated outputs. These aspects should be explicitly represented within the JSON structure. (2) \emph{Simplicity}: Strive to minimize complexity by avoiding unnecessarily nested JSON structures whenever possible.

\section{Datasets of Information Extraction Tasks}
\label{section:IE_datasets}
Table \ref{tab:IE} reports the training and evaluation datasets of information extraction tasks.

\begin{table*}[ht]
    \centering
    \begin{tabular}{ccc}
    \toprule  
    \textbf{Task} & \textbf{Dataset} & \textbf{$|$Schema$|$} \\
    \midrule
    \multicolumn{3}{c}{\textbf{Training}} \\
    \midrule \
    \multirow{17}*{NER} & ACE2004 \citep{ace2005-annotation} & 7  \\
    ~ & ACE2005 \citep{ace2005-annotation} & 7  \\
    ~ & broad\_twitter\_corpus \citep{broad_twitter_corpusDATASET} & 3\\
    ~ & CoNLL2003 \citep{CoNLL03Dataset} & 4 \\
    ~ & multiNERD \citep{multiNERDDATASET} & 16  \\
    ~ & Ontonotes \citep{OntoNotesDataset} & 18 \\
    ~ & polyglot-NER \citep{polyglot-NERDATASET} & 3 \\
    ~ & tweetNER7 \citep{tweetNER7DATASET} & 7 \\
    ~ & wikiann \citep{wikiannDataset} & 3  \\
    ~ & wikineural \citep{wikineuralDATASET} & 3  \\
    ~ & AnatEM \citep{AnatEM} & 1 \\
    ~ & bc2gm \citep{Kocaman2020BiomedicalNE} & 1  \\
    ~ & bc4chemd \citep{bc4chemdDATASET} & 1 \\
    ~ & bc5cd \citep{Li2016BioCreativeVC} & 2  \\
    ~ & FabNER \citep{Kumar2021FabNERIE} & 12 \\
    ~ & FindVehicle \citep{FindVehicle} & 21  \\
    ~ & HarveyNER \citep{HarveyNERDATASET} & 4  \\
    ~ & ncbi-disease \citep{ncbi-diseaseDATASET} & 1  \\
    \midrule
    \multirow{6}*{RE} & GIDS \citep{Jat2018ImprovingDS} & 4  \\
    ~ & kbp37 \citep{kbp37DATASET} & 18  \\
    ~ & NYT \citep{Riedel2010ModelingRA} & 24  \\
    ~ & NYT11 HRL \citep{Takanobu2018AHF} & 12 \\
    ~ & SciERC \citep{SciERCDATASET} & 7 \\
    ~ & semeval RE \citep{Hendrickx2010SemEval2010T8} & 10  \\
    \midrule
    \multicolumn{3}{c}{\textbf{Evaluation}} \\
    \midrule
    \multirow{5}*{NER}  & CrossNER\_AI \citep{CrossNER} & 14 \\
    ~ & CrossNER\_literature \citep{CrossNER} & 12  \\
    ~ & CrossNER\_music \citep{CrossNER} & 13 \\
    ~ & CrossNER\_politics \citep{CrossNER} & 9  \\
    ~ & CrossNER\_science \citep{CrossNER} & 17  \\
    \midrule
    \multirow{2}*{RE} & CoNLL2004 \citep{conll04} & 5 \\
    ~ & FewRel \citep{FewRel} & 5 \\
    \midrule
    \multirow{3}*{EE} & ACE2005 \citep{ace2005-annotation} & 33(22)  \\
    ~ & CASIE \citep{Satyapanich_Ferraro_Finin_2020} & 5(26)  \\
    ~ & PHEE \citep{PHEE} & 2(16) \\
    \bottomrule
    \end{tabular}
    \caption{Information Extraction (IE) datasets utilized for training and evaluation. $|$Schema$|$ represents the number of entity categories in the named entity recognition (NER) task, the number of relations in the relation extraction (RE) task, and the number of event categories (outside the parenthesis) along with the number of argument categories (inside the parenthesis) in the event extraction (EE) task.}
    \label{tab:IE}
\end{table*}

\section{Additional case studies on controllability}
\label{section:control}

Table \ref{tab:control2} presents additional case studies on controllability.

\begin{table*}[ht]
\begin{center}
\resizebox{\textwidth}{!}{
\begin{tabular}{p{0.08\linewidth}  p{0.7\linewidth} p{0.28\linewidth}}
 \textbf{Method} & \textbf{Input} & \textbf{Prediction}  \\ 
\toprule
\multicolumn{3}{c}{\textbf{Task: Translate a sentence to multiple languages}}  \\
\midrule
Text & Sentence: How are you? Translate the sentence to multiple languages: French and Spanish. Translations:  & Comment allez-vous?\\ 
\midrule
Json &  \{\jsonkey{``input''}: \{ 
        \jsonkey{``sentence''}: \jsonvalue{``How are you?''},
        \jsonkey{``instruction''}: \jsonvalue{``Sentence: \{sentence\}. Translate the sentence to two languages: French and Spanish. Translations: \{translations\}''}
    \},
    
    \jsonkey{``output control''}: \{
        \jsonkey{``translations''}: \{
            \jsonkey{``type''}: \jsonvalue{``object''},
            \jsonkey{``properties''}: \{
                \jsonkey{``French''}: \{
                    \jsonkey{``type''}: \jsonvalue{``string''},
                    \jsonkey{``description''}: \jsonvalue{``The French translation of the sentence.''}
                \},
                \jsonkey{``Spanish''}: \{
                    \jsonkey{``type''}: \jsonvalue{``string''},
                    \jsonkey{``description''}: \jsonvalue{``The Spanish translation of the sentence.''}
                \}
            \}
        \}
    \}
\}  & \{\jsonkey{``translations''}: \{\jsonkey{``French''}: \jsonvalue{``Comment allez-vous?''}, \jsonkey{``Spanish''}: \jsonvalue{``Cómo estás?''}\} \}\\
\midrule
\multicolumn{3}{c}{\textbf{Task: Joke generation with humor style}}  \\
\midrule
Text &  Generate the joke with a specific humor style (e.g., pun, sarcasm): & What do you call a deer with no eyes? No idea.
 \\ 
\midrule
Json & \{
    \jsonkey{``input''}: \{
        \jsonkey{``instruction''}: \jsonvalue{``Generate the joke with a specific humor style (e.g., pun, sarcasm): \{humor style\} \{joke\}}''
    \},
    
    \jsonkey{``output control''}: \{
        \jsonkey{``humor style''}: \{\jsonkey{``type''}: \jsonvalue{``string''}\},
        \jsonkey{``joke''}: \{\jsonkey{``type''}: \jsonvalue{``string''}\}
    \}
\}  & \{\jsonkey{``humor style''}: \jsonvalue{``pun''}, \jsonkey{``joke''}: \jsonvalue{``What did the pirate say when he was given a piece of paper?''}\} \\
\midrule
\multicolumn{3}{c}{\textbf{Task: Sentiment analysis with emotion}}  \\
\midrule
Text &  Given the sentiment labels: positive, negative, and neutral, and text: I just won a lottery! Life is amazing!, provide the sentiment label and emotion associated with the text. Sentiment label and emotion: & positive
 \\ 
\midrule
Json &  \{
    \jsonkey{``input''}: \{
        \jsonkey{``text''}: \jsonvalue{``I just won a lottery! Life is amazing!''},
        \jsonkey{``sentiment labels''}: \jsonvalue{[``positive'', ``negative'', ``neutral'']},
        \jsonkey{``instruction''}: \jsonvalue{``Given the sentiment labels: \{sentiment labels\} and text: \{text\}, provide the sentiment label and emotion associated with the text. Sentiment label and emotion: \{sentiment label\} \{emotion\}''}
    \},
    
    \jsonkey{``output control''}: \{
        \jsonkey{``sentiment label''}: \{\jsonkey{``type''}: \jsonvalue{``string''}\},
        \jsonkey{``emotion''}: \{\jsonkey{``type''}: \jsonvalue{``string''}\}
    \}
\} & \{\jsonkey{``sentiment label''}: \jsonvalue{``positive''}, \jsonkey{``emotion''}: \jsonvalue{``joy''}\} \\

\bottomrule
\end{tabular}
}
\end{center}
\caption{Case studies focusing on controllability. Each example displays its input along with the model's prediction.}
\label{tab:control2}
\end{table*}

\section{Introduction to JSON and Its Utilization}
\label{section:json}

\subsection{JSON Data Types and Syntax}

JSON data is represented using a combination of the following data types:

\begin{itemize}[leftmargin=0.4cm]
\item \textbf{Object:} An unordered collection of key-value pairs, enclosed in curly braces \{\}. The keys are strings, and the values can be any of the JSON data types. 
\item \textbf{Array:} An ordered list of values, enclosed in square brackets []. The values can be any of the JSON data types. \item \textbf{String:} A sequence of Unicode characters, enclosed in double quotes. 
\item \textbf{Number:} A numeric value, which can be an integer or a floating-point number.
\item \textbf{Boolean:} A value that is either true or false. 
\item \textbf{Null:} A special keyword denoting a null value. 
\end{itemize}

In this paper, we focus on the \texttt{object}, \texttt{array}, and \texttt{string} types, as the \texttt{number} and \texttt{boolean} types can be represented as the \texttt{string} type for simplicity. By combining these simple data types, JSON can represent various structured data.  This flexibility allows language models that understand basic data types to potentially generalize to more complex structures.

\subsection{JSON Schema}
JSON Schema employs a JSON-based format for defining the structure of JSON data, specifying properties like data types, required fields, and permissible values for JSON objects. It uses many keywords to define and validate JSON data. In this paper, we use the following keywords to construct the control information $C$:

\begin{itemize}[leftmargin=0.4cm]
\item \textbf{type:} Specifies the data type of a JSON value, such as \texttt{object}, \texttt{array}, and \texttt{string}. 
\item \textbf{description:} Provides explanations and clarifications about the purpose and constraints of a specific element or property. 
\item \textbf{items:} Defines the elements of an array and their data types. 
\item \textbf{properties:} Describes the properties of an object, including their data types and constraints.  
\end{itemize}

We may introduce more keywords to further improve the model's controllability in the future.

\subsection{JSON Example}

\begin{lstlisting}[language=json,firstnumber=1]
{
    "type": "object",
    "properties": {
        "first name": { "type": "string" },
        "last name": { "type": "string" },
        "phone numbers": {
            "type": "array",
            "items": { "type": "string" }
        }
        "address": {
            "type": "object",
            "properties": {
                "city": { "type": "string" },
                "state": { "type": "string" },
                "country": { "type": "string" }
            }
        }
    }
}
\end{lstlisting}

The example provided above employs JSON Schema to define a person object. This object comprises multiple properties, each with its own type. JSON's ability to handle nested structures allows it to support a wide range of complex and diverse structures. An instance of the person object can be seen below:

\begin{lstlisting}[language=json,firstnumber=1]
{
    "first name": "John",
    "last name": "Doe",
    "phone numbers": ["12345", "678910"],
    "address": {
        "city": "AnyCity",
        "state": "AnyState",
        "country": "AnyCountry"
    }
}
\end{lstlisting}

\section{Named Entity Recognition and Relation Extraction Training Prompts}
\label{section:IE_prompts}
We create prompts for both the named entity recognition (NER) and relation extraction (RE) tasks, as shown in Figures \ref{fig:ner_prompts}-\ref{fig:re_prompts_without}. For the RE task, we develop two sets of prompts: one for datasets with entity categories and another for datasets without entity categories. Each prompt comprises an input template and an output template, which are highlighted in \blue{blue} and \orange{orange}, respectively.

\begin{figure*}[ht]
\begin{center}
    \begin{mybox}
    \textcolor{red}{Prompt 1:} [\blue{definition: \{definition\}\myslash text: \{text\}\myslash entity categories: \{entity categories\}\myslash entities:}, \orange{\{entities\}}] \\
    \textcolor{red}{Prompt 2:} [\blue{definition: \{definition\}\myslash entity categories: \{entity categories\}\myslash text: \{text\}\myslash entities:}, \orange{\{entities\}}] \\
    \textcolor{red}{Prompt 3:} [\blue{text: \{text\}\myslash definition: \{definition\}\myslash entity categories: \{entity categories\}\myslash entities:}, \orange{\{entities\}}] \\
    \textcolor{red}{Prompt 4:} [\blue{text: \{text\}\myslash entity categories: \{entity categories\}\myslash definition: \{definition\}\myslash entities:}, \orange{\{entities\}}] \\
    \textcolor{red}{Prompt 5:} [\blue{entity categories: \{entity categories\}\myslash text: \{text\}\myslash definition: \{definition\}\myslash entities:}, \orange{\{entities\}}] \\
    \textcolor{red}{Prompt 6:} [\blue{entity categories: \{entity categories\}\myslash definition: \{definition\}\myslash text: \{text\}\myslash  entities:}, \orange{\{entities\}}] \\
    \textcolor{red}{Prompt 7:} [\blue{\{definition\}\myslash text: \{text\}\myslash entity categories: \{entity categories\}\myslash entities:}, \orange{\{entities\}}] \\
    \textcolor{red}{Prompt 8:} [\blue{\{definition\}\myslash entity categories: \{entity categories\}\myslash text: \{text\}\myslash entities:}, \orange{\{entities\}}] \\
    \textcolor{red}{Prompt 9:} [\blue{text: \{text\}\myslash entity categories: \{entity categories\}\myslash \{definition\}\myslash entities:}, \orange{\{entities\}}] \\
    \textcolor{red}{Prompt 10:} [\blue{entity categories: \{entity categories\}\myslash text: \{text\}\myslash \{definition\}\myslash entities:}, \orange{\{entities\}}] \\
    \end{mybox}
\end{center}
\caption{NER training prompts.}
\label{fig:ner_prompts}
\end{figure*}

\begin{figure*}[ht]
\begin{center}
    \begin{mybox}
    \textcolor{red}{Prompt 1:} [\blue{definition: \{definition\}\myslash text: \{text\}\myslash entity categories: \{entity categories\}\myslash relations: \{relations\}\myslash relational triplets:}, \orange{\{relational triplets\}}] \\
    \textcolor{red}{Prompt 2:} [\blue{definition: \{definition\}\myslash entity categories: \{entity categories\}\myslash relations: \{relations\}\myslash text: \{text\}\myslash relational triplets:}, \orange{\{relational triplets\}}] \\
    \textcolor{red}{Prompt 3:} [\blue{definition: \{definition\}\myslash text: \{text\}\myslash relations: \{relations\}\myslash entity categories: \{entity categories\}\myslash relational triplets:}, \orange{\{relational triplets\}}] \\
    \textcolor{red}{Prompt 4:} [\blue{definition: \{definition\}\myslash relations: \{relations\}\myslash entity categories: \{entity categories\}\myslash text: \{text\}\myslash relational triplets:}, \orange{\{relational triplets\}}] \\
    \textcolor{red}{Prompt 5:} [\blue{text: \{text\}\myslash definition: \{definition\}\myslash entity categories: \{entity categories\}\myslash relations: \{relations\}\myslash relational triplets:}, \orange{\{relational triplets\}}] \\
    \textcolor{red}{Prompt 6:} [\blue{text: \{text\}\myslash entity categories: \{entity categories\}\myslash relations: \{relations\}\myslash definition: \{definition\}\myslash  relational triplets:}, \orange{\{relational triplets\}}] \\
    \textcolor{red}{Prompt 7:} [\blue{text: \{text\}\myslash definition: \{definition\}\myslash relations: \{relations\} \myslash entity categories: \{entity categories\}\myslash relational triplets:}, \orange{\{relational triplets\}}] \\
    \textcolor{red}{Prompt 8:} [\blue{text: \{text\}\myslash relations: \{relations\}\myslash entity categories: \{entity categories\}\myslash definition: \{definition\}\myslash  relational triplets:}, \orange{\{relational triplets\}}] \\
    \textcolor{red}{Prompt 9:} [\blue{entity categories: \{entity categories\}\myslash relations: \{relations\}\myslash text: \{text\}\myslash definition: \{definition\}\myslash relational triplets:}, \orange{\{relational triplets\}}] \\
    \textcolor{red}{Prompt 10:} [\blue{relations: \{relations\}\myslash entity categories: \{entity categories\}\myslash definition: \{definition\}\myslash text: \{text\}\myslash relational triplets:}, \orange{\{relational triplets\}}] \\
    \end{mybox}
\end{center}
\caption{RE (with entity categories) training prompts.}
\label{fig:re_prompts_with}
\end{figure*}

\begin{figure*}[ht]
\begin{center}
    \begin{mybox}
    \textcolor{red}{Prompt 1:} [\blue{definition: \{definition\}\myslash text: \{text\}\myslash relations: \{relations\}\myslash relational triplets:}, \orange{\{relational triplets\}}] \\
    \textcolor{red}{Prompt 2:} [\blue{definition: \{definition\}\myslash relations: \{relations\}\myslash text: \{text\}\myslash relational triplets:}, \orange{\{relational triplets\}}] \\
    \textcolor{red}{Prompt 3:} [\blue{text: \{text\}\myslash definition: \{definition\}\myslash relations: \{relations\}\myslash relational triplets:}, \orange{\{relational triplets\}}] \\
    \textcolor{red}{Prompt 4:} [\blue{text: \{text\}\myslash relations: \{relations\}\myslash definition: \{definition\}\myslash relational triplets:}, \orange{\{relational triplets\}}] \\
    \textcolor{red}{Prompt 5:} [\blue{relations: \{relations\}\myslash text: \{text\}\myslash definition: \{definition\}\myslash relational triplets:}, \orange{\{relational triplets\}}] \\
    \textcolor{red}{Prompt 6:} [\blue{relations: \{relations\}\myslash definition: \{definition\}\myslash text: \{text\}\myslash  relational triplets:}, \orange{\{relational triplets\}}] \\
    \textcolor{red}{Prompt 7:} [\blue{\{definition\}\myslash text: \{text\}\myslash relations: \{relations\}\myslash relational triplets:}, \orange{\{relational triplets\}}] \\
    \textcolor{red}{Prompt 8:} [\blue{\{definition\}\myslash relations: \{relations\}\myslash text: \{text\}\myslash relational triplets:}, \orange{\{relational triplets\}}] \\
    \textcolor{red}{Prompt 9:} [\blue{text: \{text\}\myslash relations: \{relations\}\myslash \{definition\}\myslash relational triplets:}, \orange{\{relational triplets\}}] \\
    \textcolor{red}{Prompt 10:} [\blue{relations: \{relations\}\myslash text: \{text\}\myslash \{definition\}\myslash relational triplets:}, \orange{\{relational triplets\}}] \\
    \end{mybox}
\end{center}
\caption{RE (without entity categories) training prompts.}
\label{fig:re_prompts_without}
\end{figure*}

\section{Evaluation Prompts and Examples}
\label{section:evaluation}
For MMLU and BBH, we utilize the prompts from the Flan2022 collection designed for question answering\footnote{For more details, see \url{https://github.com/google-research/FLAN/blob/main/flan/v2/templates.py}.}. For other evaluation tasks, we create prompts based on their respective task components and definitions. Each prompt includes an input template and an output template, highlighted in \blue{blue} and \orange{orange}, respectively. Further details can be found in the subsequent sections.

\subsection{Generalization}
All tasks employ a single prompt for evaluation, except for the RE task. The RE task utilizes two prompts: one for datasets with entity categories and another for datasets without entity categories. The prompts and examples are presented in Figures \ref{fig:mmlu_prompt_examples}-\ref{fig:alpacaeval_prompt_examples}.

\begin{figure*}[ht]
\begin{center}
    \begin{mybox}
    \textcolor{red}{Prompt:} [\blue{\{question\}\myslash \{options\_\}\myslash Answer:}, \orange{\{answer\}}] \\
    \\
    \textcolor{red}{TextTuning Example: } \\
    \brown{Input:} The following is a multiple choice question about global facts.\myslash Controlling for inflation and PPP-adjustment, about how much did GDP per capita increase from 1950 to 2016 in Japan? Options:\myslash(A) by 5 fold\myslash(B) by 10 fold\myslash(C) by 15 fold\myslash(D) by 20 fold\myslash Answer: \\
    \brown{Output:} (C) \\
    \\
    \textcolor{red}{JsonTuning Example:} \\
    \brown{Input}: \{\jsonkey{``input''}: \{
        \jsonkey{``question''}: ``The following is a multiple choice question about global facts.\myslash Controlling for inflation and PPP-adjustment, about how much did GDP per capita increase from 1950 to 2016 in Japan?'',
        \jsonkey{``options\_''}: ``Options:\myslash(A) by 5 fold\myslash(B) by 10 fold\myslash(C) by 15 fold\myslash(D) by 20 fold'',
        \jsonkey{``instruction''}: ``\{question\}\myslash \{options\_\}\myslash Answer: \{answer\}''
    \},
    \jsonkey{``output control''}: \{
        \jsonkey{``answer''}: \{
            \jsonkey{``type''}: ``string''
        \}
    \}
\}  \\
    \brown{Output}: \{\jsonkey{``answer''}: ``(C)''\} 
    \end{mybox}
\end{center}
\caption{MMLU evaluation prompt and examples.}
\label{fig:mmlu_prompt_examples}
\end{figure*}

\begin{figure*}[ht]
\begin{center}
    \begin{mybox}
    \textcolor{red}{Prompt:} [\blue{Q: \{question\}\myslash A:}, \orange{\{answer\}}] \\
    \\
    \textcolor{red}{TextTuning Example:} \\
    \brown{Input}: Q: ((-1 + 2 + 9 * 5) - (-2 + -4 + -4 * -7)) =\myslash A: \\
    \brown{Output}: 24 \\
    \\
    \textcolor{red}{JsonTuning Example:} \\
    \brown{Input}: \{\jsonkey{``input''}: \{
        \jsonkey{``question''}: ``((-1 + 2 + 9 * 5) - (-2 + -4 + -4 * -7)) ='',
        \jsonkey{``instruction''}: ``Q: \{question\}\myslash A: \{answer\}''
    \},
    \jsonkey{``output control''}: \{
        \jsonkey{``answer''}: \{
            \jsonkey{``type''}: ``string''
        \}
    \}
\}  \\
    \brown{Output}: \{\jsonkey{``answer''}: ``24''\} 
    \end{mybox}
\end{center}
\caption{BBH evaluation prompt and examples.}
\label{fig:bbh_prompt_examples}
\end{figure*}

\begin{figure*}[ht]
\begin{center}
    \begin{mybox}
    \textcolor{red}{Prompt:} [\blue{definition: \{definition\}\myslash text: \{text\}\myslash entity categories: \{entity categories\}\myslash entities:}, \orange{\{entities\}}] \\
    \\
    \textcolor{red}{TextTuning Example:} \\
    \brown{Input}: definition: Given a text and entity categories, your task is to scan the text and identify a list of named entities in it. Each entity contains an entity category and an entity span. An entity span refers to the specific portion of the text that represents an entity. An entity category refers to the category to which an entity belongs.\myslash text: He also co-wrote Posible, which has been used as a theme song for the 2005 Southeast Asian Games.\myslash entity categories: location, event, country, band, person, song, musical artist, music genre, else, album, organization, award, musical instrument\myslash entities: \\
    \brown{Output}: [[``song'', ``Posible''], [``event'', ``2005 Southeast Asian Games'']] \\
    \\
    \textcolor{red}{JsonTuning Example:} \\
    \brown{Input}: \{\jsonkey{``input''}: \{
        \jsonkey{``definition''}: ``Given a text and entity categories, your task is to scan the text and identify a list of named entities in it. Each entity contains an entity category and an entity span. An entity span refers to the specific portion of the text that represents an entity. An entity category refers to the category to which an entity belongs.'',
        \jsonkey{``text''}: ``He also co-wrote Posible, which has been used as a theme song for the 2005 Southeast Asian Games.'',
        \jsonkey{``entity categories''}: [
            ``location'',
            ``event'',
            ``country'',
            ``band'',
            ``person'',
            ``song'',
            ``musical artist'',
            ``music genre'',
            ``else'',
            ``album'',
            ``organization'',
            ``award'',
            ``musical instrument''
        ],
        \jsonkey{``instruction''}: ``definition: \{definition\}\myslash text: \{text\}\myslash entity categories: \{entity categories\}\myslash entities: \{entities\}'',
    \},
    \jsonkey{``output control''}: \{
        \jsonkey{``entities''}: \{
            \jsonkey{``type''}: ``array'',
            \jsonkey{``items''}: \{
                \jsonkey{``type''}: ``object'',
                \jsonkey{``properties''}: \{
                    \jsonkey{``entity category''}: \{
                        \jsonkey{``type''}: ``string'',
                        \jsonkey{``description''}: ``The entity category should be one of the entity categories provided in the input.''
                    \},
                    \jsonkey{``entity span''}: \{
                        \jsonkey{``type''}: ``string'',
                        \jsonkey{``description''}: ``The entity span should be a continuous span in the text provided in the input.''
                    \}
                \}
            \}
        \}
    \}
\}    
  \\
    \brown{Output}: \{
                \jsonkey{``entities''}: [
                    \{
                        \jsonkey{``entity category''}: ``song'',
                        \jsonkey{``entity span''}: ``Posible''
                    \},
                    \{
                        \jsonkey{``entity category''}: ``event'',
                       \jsonkey{``entity span''}: ``2005 Southeast Asian Games''
                    \}
                ]
           \}
    \end{mybox}
\end{center}
\caption{NER evaluation prompt and examples.}
\label{fig:ner_prompt_examples}
\end{figure*}

\begin{figure*}[ht]
\begin{center}
    \begin{mybox}
    \textcolor{red}{Prompt:} [\blue{definition: \{definition\}\myslash text: \{text\}\myslash entity categories: \{entity categories\}\myslash relations: \{relations\}\myslash relational triplets:}, \orange{\{relational triplets\}}] \\
    \\
    \textcolor{red}{TextTuning Example:} \\
    \brown{Input}: definition: Given a text, entity categories, and relations, your goal is to scan the text and identify a list of relational triplets in it. Each relational triplet contains a head entity category, a head entity span, a relation, a tail entity category, and a tail entity span. The head entity is the subject from which the relation originates. The relation represents the specific relation between the head entity and the tail entity. The tail entity is the object which the relation points. An entity span refers to the specific portion of the text that represents an entity. An entity category refers to the category to which an entity belongs.\myslash text: In 1822, the 19th president of the United States, Rutherford B. Hayes, was born in Delaware, Ohio. \myslash entity categories: Organization, Location, People\myslash relations: Kill, Work for, Located in, Live in, Organization based in\myslash relational triplets:\\
    \brown{Output}: [[``People'', ``Rutherford B. Hayes'', ``Live in'', ``Location'', ``Delaware, Ohio'']] \\
    \\
    \textcolor{red}{JsonTuning Example:} \\
    \brown{Input}: \{\jsonkey{``input''}: \{
        \jsonkey{``definition''}: ``Given a text, entity categories, and relations, your goal is to scan the text and identify a list of relational triplets in it. Each relational triplet contains a head entity category, a head entity span, a relation, a tail entity category, and a tail entity span. The head entity is the subject from which the relation originates. The relation represents the specific relation between the head entity and the tail entity. The tail entity is the object which the relation points. An entity span refers to the specific portion of the text that represents an entity. An entity category refers to the category to which an entity belongs.'',
        \jsonkey{``text''}: ``In 1822, the 19th president of the United States, Rutherford B. Hayes, was born in Delaware, Ohio.'',
        \jsonkey{``entity categories''}: [
            ``Organization'',
            ``Location'',
            ``People''
        ],
        \jsonkey{``relations''}: [
            ``Kill'',
            ``Work for'',
            ``Located in'',
            ``Live in'',
            ``Organization based in''
        ],
        \jsonkey{``instruction''}: ``definition: \{definition\}\myslash text: \{text\}\myslash entity categories: \{entity categories\}\myslash relations: \{relations\}\myslash relational triplets: \{relational triplets\}''
            \},
    \jsonkey{``output control''}: \{
        \jsonkey{``relational triplets''}: \{
            \jsonkey{``type''}: ``array'',
            \jsonkey{``items''}: \{
                \jsonkey{``type''}: ``object'',
                \jsonkey{``properties''}: \{
                    \jsonkey{``head entity category''}: \{
                        \jsonkey{``type''}: ``string'',
                        \jsonkey{``description''}: ``The head entity category should be one of the entity categories provided in the input.''
                    \},
                    \jsonkey{``head entity span''}: \{
                        \jsonkey{``type''}: ``string'',
                        \jsonkey{``description''}: ``The head entity span should be a continuous span in the text provided in the input.''
                    \},
                    \jsonkey{``relation''}: \{
                        \jsonkey{``type''}: ``string'',
                        ``description'': ``The relation should be one of the relations provided in the input.''
                    \},
                    \jsonkey{``tail entity category''}: \{
                        \jsonkey{``type''}: ``string'',
                        \jsonkey{``description''}: ``The tail entity category should be one of the entity categories provided in the input.''
                    \},
                    \jsonkey{``tail entity span''}: \{
                        \jsonkey{``type''}: ``string'',
                        \jsonkey{``description''}: ``The tail entity span should be a continuous span in the text provided in the input.''
                    \}
                \}
            \}
        \}
    \}
            \\
    \brown{Output}: \{
                \jsonkey{``relational triplets''}: [
                    \{
                       \jsonkey{``head entity category''}: ``People'',
                        \jsonkey{``head entity span''}: ``Rutherford B. Hayes'',
                        \jsonkey{``relation''}: ``Live in'',
                        \jsonkey{``tail entity category''}: ``Location'',
                        \jsonkey{``tail entity span''}: ``Delaware, Ohio''
                    \}
                ]
            \}
    \end{mybox}
\end{center}
\caption{RE (with entity categories) evaluation prompt and examples.}
\label{fig:re_prompt_examples_with}
\end{figure*}

\begin{figure*}[ht]
\begin{center}
    \begin{mybox}
    \textcolor{red}{Prompt:} [\blue{definition: \{definition\}\myslash text: \{text\}\myslash relations: \{relations\}\myslash relational triplets:}, \orange{\{relational triplets\}}] \\
    \\
    \textcolor{red}{TextTuning Example:} \\
    \brown{Input}: definition: Given a text and relations, you are required to scan the text and identify a list of relational triplets in it. Each relational triplet contains a head entity span, a relation, and a tail entity span. The head entity is the subject from which the relation originates. The relation represents the specific relation between the head entity and the tail entity. The tail entity is the object which the relation points. An entity span refers to the specific portion of the text that represents an entity.\myslash text: The Peasants is a novel written by Nobel Prize-winning Polish author Wadysaw Reymont in four parts between 1904 and 1909.\myslash relations: place served by transport hub, winner, field of work, location of formation, occupant\myslash relational triplets:\\
    \brown{Output}: [[``Nobel Prize'', ``winner'', ``Wadysaw Reymont'']] \\
    \\
    \textcolor{red}{JsonTuning Example:} \\
    \brown{Input}: \{\jsonkey{``input''}: \{
        \jsonkey{``definition''}: ``Given a text and relations, you are required to scan the text and identify a list of relational triplets in it. Each relational triplet contains a head entity span, a relation, and a tail entity span. The head entity is the subject from which the relation originates. The relation represents the specific relation between the head entity and the tail entity. The tail entity is the object which the relation points. An entity span refers to the specific portion of the text that represents an entity.'',
        \jsonkey{``text''}: ``The Peasants is a novel written by Nobel Prize-winning Polish author Wadysaw Reymont in four parts between 1904 and 1909.'',
        \jsonkey{``relations''}: [
            ``place served by transport hub'',
            ``winner'',
            ``field of work'',
            ``location of formation'',
            ``occupant''
        ],
        \jsonkey{``instruction''}: ``\{definition\}\myslash text: \{text\}\myslash relations: \{relations\}\myslash relational triplets: \{relational triplets\}''
    \},
    \jsonkey{``output control''}: \{
        \jsonkey{``relational triplets''}: \{
            \jsonkey{``type''}: ``array'',
            \jsonkey{``items''}: \{
                \jsonkey{``type''}: ``object'',
                \jsonkey{``properties''}: \{
                    \jsonkey{``head entity span''}: \{
                        \jsonkey{``type''}: ``string'',
                        \jsonkey{``description''}: ``The head entity span should be a continuous span in the text provided in the input.''
                    \},
                    \jsonkey{``relation''}: \{
                        \jsonkey{``type''}: ``string'',
                        ``description'': ``The relation should be one of the relations provided in the input.''
                    \},
                    \jsonkey{``tail entity span''}: \{
                        \jsonkey{``type''}: ``string'',
                        \jsonkey{``description''}: ``The tail entity span should be a continuous span in the text provided in the input.''
                    \}
                \}
            \}
        \}
    \}  
             \\
    \brown{Output}: \{
                \jsonkey{``relational triplets''}: [
                    \{
                        \jsonkey{``head entity span''}: ``Nobel Prize'',
                        \jsonkey{``relation''}: ``winner'',
                        \jsonkey{``tail entity span''}: ``Wadysaw Reymont''
                    \}
                ]
            \}
    \end{mybox}
\end{center}
\caption{RE (without entity categories) evaluation prompt and examples.}
\label{fig:re_prompt_examples_without}
\end{figure*}

\begin{figure*}[ht]
\begin{center}
    \begin{mybox}
    \textcolor{red}{Prompt:} [\blue{definition: \{definition\}\myslash text: \{text\}\myslash event categories: \{event categories\}\myslash argument categories: \{argument categories\}\myslash events:}, \orange{\{events\}}] \\
    \\
    \textcolor{red}{TextTuning Example:} \\
    \brown{Input}: definition: Given a text, event categories, and argument categories, you are expected to scan the text and identify a list of events in it. Each event contains an event category, an event trigger, and a list of arguments. Each argument contains an argument category and an argument span. An event category represents the type of an event. An event trigger is the word or phrase in the text that explicitly denotes the occurrence of an event. Arguments are entities associated with an event and play specific roles or functions in relation to the event. An argument span refers to the specific portion of the text that represents an argument. An argument category refers to the category to which an argument belongs.\myslash text: Until Basra, U.S. and British troops had encountered little resistance, even when they seized nearby Umm Qasr, and moved to secure key oil fields.\myslash event categories: transfer money, start organization, extradite, meet, appeal, attack, convict, born, execute, transport, release parole, merge organization, sentence, divorce, end position, end organization, transfer ownership, start position, injure, sue, die, trial hearing, marry, nominate, charge indict, elect, declare bankruptcy, phone write, acquit, arrest jail, pardon, demonstrate, fine\myslash argument categories: instrument, vehicle, agent, seller, place, beneficiary, organization, destination, plaintiff, person, giver, recipient, victim, target, defendant, origin, prosecutor, entity, attacker, artifact, buyer, adjudicator\myslash events:'',
\\
    \brown{Output}: [[``attack'', ``seized'', [[``attacker'', ``troops''], [``place'', ``Umm Qasr'']]]]\\
    \\
    \textcolor{red}{JsonTuning Example:} \\
    \brown{Input}: \{\jsonkey{``input''}: \{
                \jsonkey{``definition''}: ``Given a text, event categories, and argument categories, you are expected to scan the text and identify a list of events in it. Each event contains an event category, an event trigger, and a list of arguments. Each argument contains an argument category and an argument span. An event category represents the type of an event. An event trigger is the word or phrase in the text that explicitly denotes the occurrence of an event. Arguments are entities associated with an event and play specific roles or functions in relation to the event. An argument span refers to the specific portion of the text that represents an argument. An argument category refers to the category to which an argument belongs.'',
                \jsonkey{``text''}: ``Until Basra, U.S. and British troops had encountered little resistance, even when they seized nearby Umm Qasr, and moved to secure key oil fields.'',
                \jsonkey{``event categories''}: [
                    ``transfer money'',
                    ``start organization'',
                    ``extradite'',
                    ``meet'',
                    ``appeal'',
                    ``attack'',
                    ``convict'',
                    ``born'',
                    ``execute'',
                    ``transport'',
                    ``release parole'',
                    ``merge organization'',
                    ``sentence'',
                    ``divorce'',
                    ``end position'',
                    ``end organization'',
                    ``transfer ownership'',
                    ``start position'',
                    ``injure'',
                    ``sue'',
                    ``die'',
                    ``trial hearing'',
                    ``marry'',
                    ``nominate'',
                    ``charge indict'',
                    ``elect'',
                    ``declare bankruptcy'',
                    ``phone write'',
                    ``acquit'',
                    ``arrest jail'',
                    ``pardon'',
                    ``demonstrate'',
                    ``fine''
                ],
                \jsonkey{``argument categories''}: [
                    ``instrument'',
                    ``vehicle'',
                    ``agent'',
                    ``seller'',
                    ``place'',
                    ``beneficiary'',
                    ``organization'',
                    ``destination'',
                    ``plaintiff'',
                    ``person'',
                    ``giver'',
                    ``recipient'',
                    ``victim'',
                    ``target'',
                    ``defendant'',
                    ``origin'',
                    ``prosecutor'',
                    ``entity'',
                    ``attacker'',
                    ``artifact'',
                    ``buyer'',
                    ``adjudicator''
                ],
                \jsonkey{``instruction''}: ``definition: \{definition\}\myslash text: \{text\}\myslash event categories: \{event categories\}\myslash argument categories: \{argument categories\}\myslash events: \{events\}''
            \},
            \jsonkey{``output control''}: \{
                \jsonkey{``events''}: \{
                    \jsonkey{``type''}: ``array'',
                    \jsonkey{``items''}: \{
                        \jsonkey{``type''}: ``object'',
                        \jsonkey{``properties''}: \{
                            \jsonkey{``event category''}: \{
                                \jsonkey{``type''}: ``string'',
                                \jsonkey{``description''}: ``The event category should be one of the event categories provided in the input.''
                            \},
                            \jsonkey{``event trigger''}: \{
                                \jsonkey{``type''}: ``string'',
                                \jsonkey{``description''}: ``The event trigger should be a continuous span in the text provided in the input.''
                            \},
                            \jsonkey{``arguments''}: \{
                                \jsonkey{``type''}: ``array'',
                                \jsonkey{``items''}: \{
                                    \jsonkey{``type''}: ``object'',
                                    \jsonkey{``properties''}: \{
                                        \jsonkey{``argument category''}: \{
                                            \jsonkey{``type''}: ``string'',
                                            \jsonkey{``description''}: ``The argument category should be one of the argument categories provided in the input.''
                                        \},
                                        \jsonkey{``argument span''}: \{
                                            \jsonkey{``type''}: ``string'',
                                            \jsonkey{``description''}: ``The argument span should be a continuous span in the text provided in the input.''
                                        \}
                                    \}
                                \}
                            \}
                        \}
                    \}
                \}
            \}
       \}     
            \\
    \brown{Output}: \{
                \jsonkey{``events''}: [
                    \{
                        \jsonkey{``event category''}: ``attack'',
                        \jsonkey{``event trigger''}: ``seized'',
                        \jsonkey{``arguments''}: [
                            \{
                                \jsonkey{``argument category''}: ``attacker'',
                                \jsonkey{``argument span''}: ``troops''
                            \},
                            \{
                                \jsonkey{``argument category''}: ``place'',
                                \jsonkey{``argument span''}: ``Umm Qasr''
                            \}
                        ]
                    \}
                ]
            \} 
    \end{mybox}
\end{center}
\caption{EE evaluation prompt and examples.}
\label{fig:ee_prompt_examples}
\end{figure*}

\begin{figure*}[ht]
\begin{center}
    \begin{mybox}
    \textcolor{red}{Prompt:} [\blue{definition: \{definition\}\myslash question: \{question\}\myslash database schema: \{database schema\}\myslash SQL query:}, \orange{\{SQL query\}}] \\
    \\
    \textcolor{red}{TextTuning Example:} \\
    \brown{Input}: definition: Given a question and database schema that consists of table names and column names in the database, the text-to-SQL parsing task aims to translate the natural language question to a sql query that can be executed on the database to produce answers.\myslash question: List the title of all cartoons in alphabetical order.\myslash database schema: Table: tv\_channel; Columns: id, series\_name, country, language, content, pixel\_aspect\_ratio\_par, hight\_definition\_tv, pay\_per\_view\_ppv, package\_option. Table: tv\_series; Columns: id, episode, air\_date, rating, share, 18\_49\_rating\_share, viewers\_m, weekly\_rank, channel. Table: cartoon; Columns: id, title, directed\_by, written\_by, original\_air\_date, production\_code, channel\myslash SQL query: \\
    \brown{Output}: select title from cartoon order by title \\
    \\
    \textcolor{red}{JsonTuning Example:} \\
    \brown{Input}: \{\jsonkey{``input''}: \{
                \jsonkey{``definition''}: ``Given a `question` and `database schema` that consists of table names and column names in the database, the text-to-SQL parsing task aims to translate the natural language question to a sql query that can be executed on the database to produce answers.'',
                \jsonkey{``question''}: ``List the title of all cartoons in alphabetical order.'',
                \jsonkey{``database schema''}: [
                    \{
                        \jsonkey{``table name''}: ``tv\_channel'',
                        \jsonkey{``column names''}: [
                            ``id'',
                            ``series\_name'',
                            ``country'',
                            ``language'',
                            ``content'',
                            ``pixel\_aspect\_ratio\_par'',
                            ``hight\_definition\_tv'',
                            ``pay\_per\_view\_ppv'',
                            ``package\_option''
                        ]
                    \},
                    \{
                        \jsonkey{``table name''}: ``tv\_series'',
                        \jsonkey{``column names''}: [
                            ``id'',
                            ``episode'',
                            ``air\_date'',
                            ``rating'',
                            ``share'',
                            ``18\_49\_rating\_share'',
                            ``viewers\_m'',
                            ``weekly\_rank'',
                            ``channel''
                        ]
                    \},
                    \{
                        \jsonkey{``table name''}: ``cartoon'',
                        \jsonkey{``column names''}: [
                            ``id'',
                            ``title'',
                            ``directed\_by'',
                            ``written\_by'',
                            ``original\_air\_date'',
                            ``production\_code'',
                            ``channel''
                        ]
                    \}
                ]
            \},
            \jsonkey{``output control''}: \{
                \jsonkey{``SQL query''}: \{
                    \jsonkey{``type''}: ``string''
                \}
            \}
        \}
            \\
    \brown{Output}: \{
                \jsonkey{``SQL query''}: ``select title from cartoon order by title''
            \}
    \end{mybox}
\end{center}
\caption{NL2SQL evaluation prompt and examples.}
\label{fig:nl2sql_prompt_examples}
\end{figure*}

\begin{figure*}[ht]
\begin{center}
    \begin{mybox}
    \textcolor{red}{Prompt:} [\blue{Q: \{question\}\myslash A:}, \orange{\{answer\}}] \\
    \\
    \textcolor{red}{TextTuning Example:} \\
    \brown{Input}: Q: Who created the Superman cartoon character?\myslash A: \\
    \brown{Output}: Superman was created by Jerry Siegel and Joe Shuster in 1938. \\
    \\
    \textcolor{red}{JsonTuning Example:} \\
    \brown{Input}: \{\jsonkey{``input''}: \{
        \jsonkey{``question''}: ``Who created the Superman cartoon character?'',
        \jsonkey{``instruction''}: ``Q: \{question\}\myslash A: \{answer\}''
    \},
    \jsonkey{``output control''}: \{
        \jsonkey{``answer''}: \{
            \jsonkey{``type''}: ``string''
        \}
    \}
\}  \\
    \brown{Output}: \{\jsonkey{``answer''}: ``Superman was created by Jerry Siegel and Joe Shuster in 1938.''\} 
    \end{mybox}
\end{center}
\caption{AlpacaEval evaluation prompt and examples.}
\label{fig:alpacaeval_prompt_examples}
\end{figure*}

\subsection{Robustness}

For evaluation, we employ 10 prompts for the MMLU benchmark, the NER task, and the NL2SQL task. Prompts are demonstrated in Figure \ref{fig:mmlu_robust_prompts}, Figure \ref{fig:ner_prompts}, and Figure \ref{fig:nl2sql_robust_prompts}, respectively.

\begin{figure*}[ht]
\begin{center}
    \begin{mybox}
    \textcolor{red}{Prompt 1:} [\blue{\{question\}\myslash \{options\_\}\myslash Answer:}, \orange{\{answer\}}] \\
    \textcolor{red}{Prompt 2:} [\blue{\{question\}\myslash\myslash \{options\_\}\myslash Answer:}, \orange{\{answer\}}] \\
    \textcolor{red}{Prompt 3:} [\blue{\{question\}\myslash \{options\_\}}, \orange{\{answer\}}] \\
    \textcolor{red}{Prompt 4:} [\blue{Q: \{question\}\myslash\myslash \{options\_\}\myslash A:}, \orange{\{answer\}}] \\
    \textcolor{red}{Prompt 5:} [\blue{Answer the following question: \{question\}\myslash\myslash \{options\_\}\myslash Answer:}, \orange{\{answer\}}] \\
    \textcolor{red}{Prompt 6:} [\blue{\{options\_\}\myslash\myslash \{question\}\myslash Answer:}, \orange{\{answer\}}] \\
    \textcolor{red}{Prompt 7:} [\blue{\{options\_\}\myslash Q: \{question\}\myslash A:}, \orange{\{answer\}}] \\
    \textcolor{red}{Prompt 8:} [\blue{\{question\}\myslash\myslash \{options\_\}\myslash The answer is:}, \orange{\{answer\}}] \\
    \textcolor{red}{Prompt 9:} [\blue{\{options\_\}\myslash Given those answer options, answer the question: \{question\}\myslash  Answer:}, \orange{\{answer\}}] \\
    \textcolor{red}{Prompt 10:} [\blue{Q: \{question\}\myslash\myslash \{options\_\}\myslash The answer is:}, \orange{\{answer\}}] \\
    \end{mybox}
\end{center}
\caption{MMLU robustness evaluation prompts.}
\label{fig:mmlu_robust_prompts}
\end{figure*}

\begin{figure*}[ht]
\begin{center}
    \begin{mybox}
    \textcolor{red}{Prompt 1:} [\blue{definition: \{definition\}\myslash question: \{question\}\myslash database schema: \{database schema\}\myslash SQL query:}, \orange{\{SQL query\}}] \\
    \textcolor{red}{Prompt 2:} [\blue{definition: \{definition\}\myslash database schema: \{database schema\}\myslash question: \{question\}\myslash SQL query:}, \orange{\{SQL query\}}] \\
    \textcolor{red}{Prompt 3:} [\blue{question: \{question\}\myslash definition: \{definition\}\myslash database schema: \{database schema\}\myslash SQL query:}, \orange{\{SQL query\}}] \\
    \textcolor{red}{Prompt 4:} [\blue{question: \{question\}\myslash database schema: \{database schema\}\myslash definition: \{definition\}\myslash SQL query:}, \orange{\{SQL query\}}] \\
    \textcolor{red}{Prompt 5:} [\blue{database schema: \{database schema\}\myslash question: \{question\}\myslash definition: \{definition\}\myslash SQL query:}, \orange{\{SQL query\}}] \\
    \textcolor{red}{Prompt 6:} [\blue{database schema: \{database schema\}\myslash definition: \{definition\}\myslash question: \{question\}\myslash  SQL query:}, \orange{\{SQL query\}}] \\
    \textcolor{red}{Prompt 7:} [\blue{\{definition\}\myslash question: \{question\}\myslash database schema: \{database schema\}\myslash SQL query:}, \orange{\{SQL query\}}] \\
    \textcolor{red}{Prompt 8:} [\blue{\{definition\}\myslash database schema: \{database schema\}\myslash question: \{question\}\myslash SQL query:}, \orange{\{SQL query\}}] \\
    \textcolor{red}{Prompt 9:} [\blue{question: \{question\}\myslash database schema: \{database schema\}\myslash \{definition\}\myslash SQL query:}, \orange{\{SQL query\}}] \\
    \textcolor{red}{Prompt 10:} [\blue{database schema: \{database schema\}\myslash question: \{question\}\myslash \{definition\}\myslash SQL query:}, \orange{\{SQL query\}}] \\
    \end{mybox}
\end{center}
\caption{NL2SQL robustness evaluation prompts.}
\label{fig:nl2sql_robust_prompts}
\end{figure*}


\begin{thebibliography}{82}
\providecommand{\natexlab}[1]{#1}

\bibitem[{Al-Rfou et~al.(2015)Al-Rfou, Kulkarni, Perozzi, and Skiena}]{polyglot-NERDATASET}
Rami Al-Rfou, Vivek Kulkarni, Bryan Perozzi, and Steven Skiena. 2015.
\newblock \href {https://arxiv.org/abs/1410.3791} {Polyglot-ner: Massive multilingual named entity recognition}.
\newblock In \emph{Proceedings of the 2015 SIAM International Conference on Data Mining}, pages 586--594.

\bibitem[{Anthropic(2024)}]{Claude3}
Anthropic. 2024.
\newblock \href {https://www.anthropic.com/news/claude-3-family} {Introducing the next generation of claude}.

\bibitem[{Bai et~al.(2023)Bai, Bai, Chu, Cui, Dang, Deng, Fan, Ge, Han, Huang et~al.}]{bai2023qwen}
Jinze Bai, Shuai Bai, Yunfei Chu, Zeyu Cui, Kai Dang, Xiaodong Deng, Yang Fan, Wenbin Ge, Yu~Han, Fei Huang, et~al. 2023.
\newblock \href {https://arxiv.org/abs/2309.16609} {Qwen technical report}.
\newblock \emph{arXiv preprint arXiv:2309.16609}.

\bibitem[{Bai et~al.(2022)Bai, Jones, Ndousse, Askell, Chen, DasSarma, Drain, Fort, Ganguli, Henighan et~al.}]{bai2022training}
Yuntao Bai, Andy Jones, Kamal Ndousse, Amanda Askell, Anna Chen, Nova DasSarma, Dawn Drain, Stanislav Fort, Deep Ganguli, Tom Henighan, et~al. 2022.
\newblock \href {https://arxiv.org/abs/2204.05862} {Training a helpful and harmless assistant with reinforcement learning from human feedback}.
\newblock \emph{arXiv preprint arXiv:2204.05862}.

\bibitem[{Brown et~al.(2020)Brown, Mann, Ryder, Subbiah, Kaplan, Dhariwal, Neelakantan, Shyam, Sastry, Askell, Agarwal, Herbert-Voss, Krueger, Henighan, Child, Ramesh, Ziegler, Wu, Winter, Hesse, Chen, Sigler, Litwin, Gray, Chess, Clark, Berner, McCandlish, Radford, Sutskever, and Amodei}]{GPT-3}
Tom Brown, Benjamin Mann, Nick Ryder, Melanie Subbiah, Jared~D Kaplan, Prafulla Dhariwal, Arvind Neelakantan, Pranav Shyam, Girish Sastry, Amanda Askell, Sandhini Agarwal, Ariel Herbert-Voss, Gretchen Krueger, Tom Henighan, Rewon Child, Aditya Ramesh, Daniel Ziegler, Jeffrey Wu, Clemens Winter, Chris Hesse, Mark Chen, Eric Sigler, Mateusz Litwin, Scott Gray, Benjamin Chess, Jack Clark, Christopher Berner, Sam McCandlish, Alec Radford, Ilya Sutskever, and Dario Amodei. 2020.
\newblock \href {https://proceedings.neurips.cc/paper_files/paper/2020/file/1457c0d6bfcb4967418bfb8ac142f64a-Paper.pdf} {Language models are few-shot learners}.
\newblock In \emph{Advances in Neural Information Processing Systems}, volume~33, pages 1877--1901. Curran Associates, Inc.

\bibitem[{Cao et~al.(2023)Cao, Kang, and Sun}]{cao2023instruction}
Yihan Cao, Yanbin Kang, and Lichao Sun. 2023.
\newblock \href {https://arxiv.org/abs/2307.06290} {Instruction mining: High-quality instruction data selection for large language models}.
\newblock \emph{arXiv preprint arXiv:2307.06290}.

\bibitem[{Chen et~al.(2022)Chen, Xu, Zhang, and Huang}]{HarveyNERDATASET}
Pei Chen, Haotian Xu, Cheng Zhang, and Ruihong Huang. 2022.
\newblock \href {https://doi.org/10.18653/v1/2022.naacl-main.243} {Crossroads, buildings and neighborhoods: A dataset for fine-grained location recognition}.
\newblock In \emph{Proceedings of the 2022 Conference of the North American Chapter of the Association for Computational Linguistics: Human Language Technologies}, pages 3329--3339, Seattle, United States. Association for Computational Linguistics.

\bibitem[{Chowdhery et~al.(2022)Chowdhery, Narang, Devlin, Bosma, Mishra, Roberts, Barham, Chung, Sutton, Gehrmann et~al.}]{PaLM}
Aakanksha Chowdhery, Sharan Narang, Jacob Devlin, Maarten Bosma, Gaurav Mishra, Adam Roberts, Paul Barham, Hyung~Won Chung, Charles Sutton, Sebastian Gehrmann, et~al. 2022.
\newblock \href {https://arxiv.org/abs/2204.02311} {Palm: Scaling language modeling with pathways}.
\newblock \emph{arXiv preprint arXiv:2204.02311}.

\bibitem[{Chung et~al.(2024)Chung, Hou, Longpre, Zoph, Tay, Fedus, Li, Wang, Dehghani, Brahma, Webson, Gu, Dai, Suzgun, Chen, Chowdhery, Castro-Ros, Pellat, Robinson, Valter, Narang, Mishra, Yu, Zhao, Huang, Dai, Yu, Petrov, Chi, Dean, Devlin, Roberts, Zhou, Le, and Wei}]{chung2022scaling}
Hyung~Won Chung, Le~Hou, Shayne Longpre, Barret Zoph, Yi~Tay, William Fedus, Yunxuan Li, Xuezhi Wang, Mostafa Dehghani, Siddhartha Brahma, Albert Webson, Shixiang~Shane Gu, Zhuyun Dai, Mirac Suzgun, Xinyun Chen, Aakanksha Chowdhery, Alex Castro-Ros, Marie Pellat, Kevin Robinson, Dasha Valter, Sharan Narang, Gaurav Mishra, Adams Yu, Vincent Zhao, Yanping Huang, Andrew Dai, Hongkun Yu, Slav Petrov, Ed~H. Chi, Jeff Dean, Jacob Devlin, Adam Roberts, Denny Zhou, Quoc~V. Le, and Jason Wei. 2024.
\newblock \href {http://jmlr.org/papers/v25/23-0870.html} {Scaling instruction-finetuned language models}.
\newblock \emph{Journal of Machine Learning Research}, 25(70):1--53.

\bibitem[{Collaboration(2023)}]{srivastava2023beyond}
BIG-Bench Collaboration. 2023.
\newblock \href {https://openreview.net/forum?id=uyTL5Bvosj} {Beyond the imitation game: Quantifying and extrapolating the capabilities of language models}.
\newblock \emph{Transactions on Machine Learning Research}.

\bibitem[{Databricks(2023)}]{dolly}
Databricks. 2023.
\newblock \href {https://www.databricks.com/blog/2023/04/12/dolly-first-open-commercially-viable-instruction-tuned-llm} {Free dolly: Introducing the world's first truly open instruction-tuned llm}.
\newblock Blog post.

\bibitem[{Derczynski et~al.(2016)Derczynski, Bontcheva, and Roberts}]{broad_twitter_corpusDATASET}
Leon Derczynski, Kalina Bontcheva, and Ian Roberts. 2016.
\newblock \href {https://aclanthology.org/C16-1111} {Broad {T}witter corpus: A diverse named entity recognition resource}.
\newblock In \emph{Proceedings of {COLING} 2016, the 26th International Conference on Computational Linguistics: Technical Papers}, pages 1169--1179, Osaka, Japan. The COLING 2016 Organizing Committee.

\bibitem[{Deutsch et~al.(2019)Deutsch, Upadhyay, and Roth}]{deutsch-etal-2019-general}
Daniel Deutsch, Shyam Upadhyay, and Dan Roth. 2019.
\newblock \href {https://doi.org/10.18653/v1/K19-1045} {A general-purpose algorithm for constrained sequential inference}.
\newblock In \emph{Proceedings of the 23rd Conference on Computational Natural Language Learning (CoNLL)}, pages 482--492, Hong Kong, China. Association for Computational Linguistics.

\bibitem[{Ding et~al.(2023)Ding, Chen, Xu, Qin, Hu, Liu, Sun, and Zhou}]{ultrachat}
Ning Ding, Yulin Chen, Bokai Xu, Yujia Qin, Shengding Hu, Zhiyuan Liu, Maosong Sun, and Bowen Zhou. 2023.
\newblock \href {https://doi.org/10.18653/v1/2023.emnlp-main.183} {Enhancing chat language models by scaling high-quality instructional conversations}.
\newblock In \emph{Proceedings of the 2023 Conference on Empirical Methods in Natural Language Processing}, pages 3029--3051, Singapore. Association for Computational Linguistics.

\bibitem[{Dogan et~al.(2014)Dogan, Leaman, and Lu}]{ncbi-diseaseDATASET}
Rezarta~Islamaj Dogan, Robert Leaman, and Zhiyong Lu. 2014.
\newblock Ncbi disease corpus: A resource for disease name recognition and concept normalization.
\newblock \emph{Journal of biomedical informatics}, 47:1--10.

\bibitem[{Dubey et~al.(2024)Dubey, Jauhri, Pandey, Kadian, Al-Dahle, Letman, Mathur, Schelten, Yang, Fan et~al.}]{llama3}
Abhimanyu Dubey, Abhinav Jauhri, Abhinav Pandey, Abhishek Kadian, Ahmad Al-Dahle, Aiesha Letman, Akhil Mathur, Alan Schelten, Amy Yang, Angela Fan, et~al. 2024.
\newblock \href {https://arxiv.org/abs/2407.21783} {The llama 3 herd of models}.
\newblock \emph{arXiv preprint arXiv:2407.21783}.

\bibitem[{Geng et~al.(2023)Geng, Josifoski, Peyrard, and West}]{geng-etal-2023-grammar}
Saibo Geng, Martin Josifoski, Maxime Peyrard, and Robert West. 2023.
\newblock \href {https://doi.org/10.18653/v1/2023.emnlp-main.674} {Grammar-constrained decoding for structured {NLP} tasks without finetuning}.
\newblock In \emph{Proceedings of the 2023 Conference on Empirical Methods in Natural Language Processing}, pages 10932--10952, Singapore. Association for Computational Linguistics.

\bibitem[{Guan et~al.(2023)Guan, Man, Chen, Yao, Hu, Zhu, Smith, Lim, and Yue}]{FindVehicle}
Runwei Guan, Ka~Lok Man, Feifan Chen, Shanliang Yao, Rongsheng Hu, Xiaohui Zhu, Jeremy Smith, Eng~Gee Lim, and Yutao Yue. 2023.
\newblock \href {https://arxiv.org/abs/2304.10893} {Findvehicle and vehiclefinder: A ner dataset for natural language-based vehicle retrieval and a keyword-based cross-modal vehicle retrieval system}.
\newblock \emph{arXiv preprint arXiv:2304.10893}.

\bibitem[{Han et~al.(2023)Han, Peng, Yang, Wang, Liu, and Wan}]{han2023information}
Ridong Han, Tao Peng, Chaohao Yang, Benyou Wang, Lu~Liu, and Xiang Wan. 2023.
\newblock \href {https://arxiv.org/abs/2305.14450} {Is information extraction solved by chatgpt? an analysis of performance, evaluation criteria, robustness and errors}.
\newblock \emph{arXiv preprint arXiv:2305.14450}.

\bibitem[{Han et~al.(2018)Han, Zhu, Yu, Wang, Yao, Liu, and Sun}]{FewRel}
Xu~Han, Hao Zhu, Pengfei Yu, Ziyun Wang, Yuan Yao, Zhiyuan Liu, and Maosong Sun. 2018.
\newblock \href {https://doi.org/10.18653/v1/D18-1514} {{F}ew{R}el: A large-scale supervised few-shot relation classification dataset with state-of-the-art evaluation}.
\newblock In \emph{Proceedings of the 2018 Conference on Empirical Methods in Natural Language Processing}, pages 4803--4809, Brussels, Belgium. Association for Computational Linguistics.

\bibitem[{Hendrickx et~al.(2010)Hendrickx, Kim, Kozareva, Nakov, {\'O}~S{\'e}aghdha, Pad{\'o}, Pennacchiotti, Romano, and Szpakowicz}]{Hendrickx2010SemEval2010T8}
Iris Hendrickx, Su~Nam Kim, Zornitsa Kozareva, Preslav Nakov, Diarmuid {\'O}~S{\'e}aghdha, Sebastian Pad{\'o}, Marco Pennacchiotti, Lorenza Romano, and Stan Szpakowicz. 2010.
\newblock \href {https://aclanthology.org/S10-1006} {{S}em{E}val-2010 task 8: Multi-way classification of semantic relations between pairs of nominals}.
\newblock In \emph{Proceedings of the 5th International Workshop on Semantic Evaluation}, pages 33--38, Uppsala, Sweden. Association for Computational Linguistics.

\bibitem[{Hendrycks et~al.(2021)Hendrycks, Burns, Basart, Zou, Mazeika, Song, and Steinhardt}]{MMLU}
Dan Hendrycks, Collin Burns, Steven Basart, Andy Zou, Mantas Mazeika, Dawn Song, and Jacob Steinhardt. 2021.
\newblock \href {https://openreview.net/forum?id=d7KBjmI3GmQ} {Measuring massive multitask language understanding}.
\newblock In \emph{International Conference on Learning Representations}.

\bibitem[{Hovy et~al.(2006)Hovy, Marcus, Palmer, Ramshaw, and Weischedel}]{OntoNotesDataset}
Eduard Hovy, Mitchell Marcus, Martha Palmer, Lance Ramshaw, and Ralph Weischedel. 2006.
\newblock \href {https://aclanthology.org/N06-2015} {{O}nto{N}otes: The 90{\%} solution}.
\newblock In \emph{Proceedings of the Human Language Technology Conference of the {NAACL}, Companion Volume: Short Papers}, pages 57--60, New York City, USA. Association for Computational Linguistics.

\bibitem[{Hu et~al.(2022)Hu, yelong shen, Wallis, Allen-Zhu, Li, Wang, Wang, and Chen}]{hu2022lora}
Edward~J Hu, yelong shen, Phillip Wallis, Zeyuan Allen-Zhu, Yuanzhi Li, Shean Wang, Lu~Wang, and Weizhu Chen. 2022.
\newblock \href {https://openreview.net/forum?id=nZeVKeeFYf9} {Lo{RA}: Low-rank adaptation of large language models}.
\newblock In \emph{International Conference on Learning Representations}.

\bibitem[{Iyer et~al.(2022)Iyer, Lin, Pasunuru, Mihaylov, Simig, Yu, Shuster, Wang, Liu, Koura et~al.}]{opt-iml}
Srinivasan Iyer, Xi~Victoria Lin, Ramakanth Pasunuru, Todor Mihaylov, Daniel Simig, Ping Yu, Kurt Shuster, Tianlu Wang, Qing Liu, Punit~Singh Koura, et~al. 2022.
\newblock \href {https://arxiv.org/abs/2212.12017} {Opt-iml: Scaling language model instruction meta learning through the lens of generalization}.
\newblock \emph{arXiv preprint arXiv:2212.12017}.

\bibitem[{Jiang et~al.(2023)Jiang, Sablayrolles, Mensch, Bamford, Chaplot, Casas, Bressand, Lengyel, Lample, Saulnier et~al.}]{mistral}
Albert~Q Jiang, Alexandre Sablayrolles, Arthur Mensch, Chris Bamford, Devendra~Singh Chaplot, Diego de~las Casas, Florian Bressand, Gianna Lengyel, Guillaume Lample, Lucile Saulnier, et~al. 2023.
\newblock \href {https://arxiv.org/abs/2310.06825} {Mistral 7b}.
\newblock \emph{arXiv preprint arXiv:2310.06825}.

\bibitem[{Kocaman and Talby(2021)}]{Kocaman2020BiomedicalNE}
Veysel Kocaman and David Talby. 2021.
\newblock Biomedical named entity recognition at scale.
\newblock In \emph{Pattern Recognition. ICPR International Workshops and Challenges}, pages 635--646, Cham. Springer International Publishing.

\bibitem[{K{\"o}pf et~al.(2023)K{\"o}pf, Kilcher, von R{\"u}tte, Anagnostidis, Tam, Stevens, Barhoum, Duc, Stanley, Nagyfi et~al.}]{kopf2023openassistant}
Andreas K{\"o}pf, Yannic Kilcher, Dimitri von R{\"u}tte, Sotiris Anagnostidis, Zhi-Rui Tam, Keith Stevens, Abdullah Barhoum, Nguyen~Minh Duc, Oliver Stanley, Rich{\'a}rd Nagyfi, et~al. 2023.
\newblock \href {https://arxiv.org/abs/2304.07327} {Openassistant conversations--democratizing large language model alignment}.
\newblock \emph{arXiv preprint arXiv:2304.07327}.

\bibitem[{Krallinger et~al.(2015)Krallinger, Rabal, Leitner, Vazquez, Salgado, lu, Leaman, Lu, Ji, Lowe, Sayle, Batista-Navarro, Rak, Huber, Rocktäschel, Matos, Campos, Tang, Qi, and Valencia}]{bc4chemdDATASET}
Martin Krallinger, Obdulia Rabal, Florian Leitner, Miguel Vazquez, David Salgado, Zhiyong lu, Robert Leaman, Yanan Lu, Donghong Ji, Daniel Lowe, Roger Sayle, Riza Batista-Navarro, Rafal Rak, Torsten Huber, Tim Rocktäschel, Sérgio Matos, David Campos, Buzhou Tang, Wang Qi, and Alfonso Valencia. 2015.
\newblock \href {https://doi.org/10.1186/1758-2946-7-S1-S2} {The chemdner corpus of chemicals and drugs and its annotation principles}.
\newblock \emph{Journal of Cheminformatics}, 7:S2.

\bibitem[{Kumar and Starly(2021)}]{Kumar2021FabNERIE}
Aman Kumar and Binil Starly. 2021.
\newblock “fabner”: information extraction from manufacturing process science domain literature using named entity recognition.
\newblock \emph{Journal of Intelligent Manufacturing}, 33:2393 -- 2407.

\bibitem[{Li et~al.(2016)Li, Sun, Johnson, Sciaky, Wei, Leaman, Davis, Mattingly, Wiegers, and Lu}]{Li2016BioCreativeVC}
Jiao Li, Yueping Sun, Robin~J. Johnson, Daniela Sciaky, Chih-Hsuan Wei, Robert Leaman, Allan~Peter Davis, Carolyn~J. Mattingly, Thomas~C. Wiegers, and Zhiyong Lu. 2016.
\newblock Biocreative v cdr task corpus: a resource for chemical disease relation extraction.
\newblock \emph{Database: The Journal of Biological Databases and Curation}, 2016.

\bibitem[{Li et~al.(2023)Li, Zhang, Dubois, Taori, Gulrajani, Guestrin, Liang, and Hashimoto}]{alpaca_eval}
Xuechen Li, Tianyi Zhang, Yann Dubois, Rohan Taori, Ishaan Gulrajani, Carlos Guestrin, Percy Liang, and Tatsunori~B. Hashimoto. 2023.
\newblock \href {https://github.com/tatsu-lab/alpaca_eval} {Alpacaeval: An automatic evaluator of instruction-following models}.
\newblock Github repository.

\bibitem[{Liu et~al.(2024)Liu, Zeng, He, Jiang, and He}]{liu2023makes}
Wei Liu, Weihao Zeng, Keqing He, Yong Jiang, and Junxian He. 2024.
\newblock \href {https://openreview.net/forum?id=BTKAeLqLMw} {What makes good data for alignment? a comprehensive study of automatic data selection in instruction tuning}.
\newblock In \emph{The Twelfth International Conference on Learning Representations}.

\bibitem[{Liu et~al.(2021)Liu, Xu, Yu, Dai, Ji, Cahyawijaya, Madotto, and Fung}]{CrossNER}
Zihan Liu, Yan Xu, Tiezheng Yu, Wenliang Dai, Ziwei Ji, Samuel Cahyawijaya, Andrea Madotto, and Pascale Fung. 2021.
\newblock \href {https://doi.org/10.1609/aaai.v35i15.17587} {Crossner: Evaluating cross-domain named entity recognition}.
\newblock \emph{Proceedings of the AAAI Conference on Artificial Intelligence}, 35(15):13452--13460.

\bibitem[{Longpre et~al.(2023)Longpre, Hou, Vu, Webson, Chung, Tay, Zhou, Le, Zoph, Wei, and Roberts}]{longpre2023flan}
Shayne Longpre, Le~Hou, Tu~Vu, Albert Webson, Hyung~Won Chung, Yi~Tay, Denny Zhou, Quoc~V Le, Barret Zoph, Jason Wei, and Adam Roberts. 2023.
\newblock \href {https://proceedings.mlr.press/v202/longpre23a.html} {The flan collection: Designing data and methods for effective instruction tuning}.
\newblock In \emph{Proceedings of the 40th International Conference on Machine Learning}, volume 202 of \emph{Proceedings of Machine Learning Research}, pages 22631--22648. PMLR.

\bibitem[{Loshchilov and Hutter(2019)}]{AdamW}
Ilya Loshchilov and Frank Hutter. 2019.
\newblock \href {https://openreview.net/forum?id=Bkg6RiCqY7} {Decoupled weight decay regularization}.
\newblock In \emph{International Conference on Learning Representations}.

\bibitem[{Lu et~al.(2024)Lu, Yuan, Yuan, Lin, Lin, Tan, Zhou, and Zhou}]{lu2023instag}
Keming Lu, Hongyi Yuan, Zheng Yuan, Runji Lin, Junyang Lin, Chuanqi Tan, Chang Zhou, and Jingren Zhou. 2024.
\newblock \href {https://openreview.net/forum?id=pszewhybU9} {\#instag: Instruction tagging for analyzing supervised fine-tuning of large language models}.
\newblock In \emph{The Twelfth International Conference on Learning Representations}.

\bibitem[{Luan et~al.(2018)Luan, He, Ostendorf, and Hajishirzi}]{SciERCDATASET}
Yi~Luan, Luheng He, Mari Ostendorf, and Hannaneh Hajishirzi. 2018.
\newblock \href {https://doi.org/10.18653/v1/D18-1360} {Multi-task identification of entities, relations, and coreference for scientific knowledge graph construction}.
\newblock In \emph{Proceedings of the 2018 Conference on Empirical Methods in Natural Language Processing}, pages 3219--3232, Brussels, Belgium. Association for Computational Linguistics.

\bibitem[{Mishra et~al.(2022)Mishra, Khashabi, Baral, and Hajishirzi}]{mishra-etal-2022-cross}
Swaroop Mishra, Daniel Khashabi, Chitta Baral, and Hannaneh Hajishirzi. 2022.
\newblock \href {https://doi.org/10.18653/v1/2022.acl-long.244} {Cross-task generalization via natural language crowdsourcing instructions}.
\newblock In \emph{Proceedings of the 60th Annual Meeting of the Association for Computational Linguistics (Volume 1: Long Papers)}, pages 3470--3487, Dublin, Ireland. Association for Computational Linguistics.

\bibitem[{Nayak et~al.(2021)Nayak, Majumder, and Poria}]{Jat2018ImprovingDS}
Tapas Nayak, Navonil Majumder, and Soujanya Poria. 2021.
\newblock \href {https://aclanthology.org/2021.ranlp-1.116} {Improving distantly supervised relation extraction with self-ensemble noise filtering}.
\newblock In \emph{Proceedings of the International Conference on Recent Advances in Natural Language Processing (RANLP 2021)}, pages 1031--1039, Held Online. INCOMA Ltd.

\bibitem[{OpenAI(2023{\natexlab{a}})}]{GPT-4}
OpenAI. 2023{\natexlab{a}}.
\newblock \href {https://arxiv.org/abs/2303.08774} {Gpt-4 technical report}.
\newblock \emph{arXiv preprint arXiv:2303.08774}.

\bibitem[{OpenAI(2023{\natexlab{b}})}]{Chatgpt}
OpenAI. 2023{\natexlab{b}}.
\newblock \href {https://openai.com/blog/chatgpt} {Introducing chatgpt}.

\bibitem[{Ouyang et~al.(2022)Ouyang, Wu, Jiang, Almeida, Wainwright, Mishkin, Zhang, Agarwal, Slama, Gray, Schulman, Hilton, Kelton, Miller, Simens, Askell, Welinder, Christiano, Leike, and Lowe}]{InstructGPT}
Long Ouyang, Jeffrey Wu, Xu~Jiang, Diogo Almeida, Carroll Wainwright, Pamela Mishkin, Chong Zhang, Sandhini Agarwal, Katarina Slama, Alex Gray, John Schulman, Jacob Hilton, Fraser Kelton, Luke Miller, Maddie Simens, Amanda Askell, Peter Welinder, Paul Christiano, Jan Leike, and Ryan Lowe. 2022.
\newblock \href {https://openreview.net/forum?id=TG8KACxEON} {Training language models to follow instructions with human feedback}.
\newblock In \emph{Advances in Neural Information Processing Systems}.

\bibitem[{Pan et~al.(2017)Pan, Zhang, May, Nothman, Knight, and Ji}]{wikiannDataset}
Xiaoman Pan, Boliang Zhang, Jonathan May, Joel Nothman, Kevin Knight, and Heng Ji. 2017.
\newblock \href {https://doi.org/10.18653/v1/P17-1178} {Cross-lingual name tagging and linking for 282 languages}.
\newblock In \emph{Proceedings of the 55th Annual Meeting of the Association for Computational Linguistics (Volume 1: Long Papers)}, pages 1946--1958, Vancouver, Canada. Association for Computational Linguistics.

\bibitem[{Paolini et~al.(2021)Paolini, Athiwaratkun, Krone, Ma, Achille, ANUBHAI, dos Santos, Xiang, and Soatto}]{TANL}
Giovanni Paolini, Ben Athiwaratkun, Jason Krone, Jie Ma, Alessandro Achille, RISHITA ANUBHAI, Cicero~Nogueira dos Santos, Bing Xiang, and Stefano Soatto. 2021.
\newblock \href {https://openreview.net/forum?id=US-TP-xnXI} {Structured prediction as translation between augmented natural languages}.
\newblock In \emph{International Conference on Learning Representations}.

\bibitem[{Penedo et~al.(2023)Penedo, Malartic, Hesslow, Cojocaru, Cappelli, Alobeidli, Pannier, Almazrouei, and Launay}]{Falcon}
Guilherme Penedo, Quentin Malartic, Daniel Hesslow, Ruxandra Cojocaru, Alessandro Cappelli, Hamza Alobeidli, Baptiste Pannier, Ebtesam Almazrouei, and Julien Launay. 2023.
\newblock \href {https://arxiv.org/abs/2306.01116} {The refinedweb dataset for falcon llm: outperforming curated corpora with web data, and web data only}.
\newblock \emph{arXiv preprint arXiv:2306.01116}.

\bibitem[{Peng et~al.(2023)Peng, Li, He, Galley, and Gao}]{peng2023instruction}
Baolin Peng, Chunyuan Li, Pengcheng He, Michel Galley, and Jianfeng Gao. 2023.
\newblock \href {https://arxiv.org/abs/2304.03277} {Instruction tuning with gpt-4}.
\newblock \emph{arXiv preprint arXiv:2304.03277}.

\bibitem[{Pezeshkpour and Hruschka(2024)}]{pezeshkpour2023large}
Pouya Pezeshkpour and Estevam Hruschka. 2024.
\newblock \href {https://doi.org/10.18653/v1/2024.findings-naacl.130} {Large language models sensitivity to the order of options in multiple-choice questions}.
\newblock In \emph{Findings of the Association for Computational Linguistics: NAACL 2024}, pages 2006--2017, Mexico City, Mexico. Association for Computational Linguistics.

\bibitem[{Pyysalo and Ananiadou(2013)}]{AnatEM}
Sampo Pyysalo and Sophia Ananiadou. 2013.
\newblock \href {https://doi.org/10.1093/bioinformatics/btt580} {{Anatomical entity mention recognition at literature scale}}.
\newblock \emph{Bioinformatics}, 30(6):868--875.

\bibitem[{Riedel et~al.(2010)Riedel, Yao, and McCallum}]{Riedel2010ModelingRA}
Sebastian Riedel, Limin Yao, and Andrew McCallum. 2010.
\newblock Modeling relations and their mentions without labeled text.
\newblock In \emph{Machine Learning and Knowledge Discovery in Databases}, pages 148--163, Berlin, Heidelberg. Springer Berlin Heidelberg.

\bibitem[{Roth and Yih(2004)}]{conll04}
Dan Roth and Wen-tau Yih. 2004.
\newblock \href {https://aclanthology.org/W04-2401} {A linear programming formulation for global inference in natural language tasks}.
\newblock In \emph{Proceedings of the Eighth Conference on Computational Natural Language Learning ({C}o{NLL}-2004) at {HLT}-{NAACL} 2004}, pages 1--8, Boston, Massachusetts, USA. Association for Computational Linguistics.

\bibitem[{Sanh et~al.(2022)Sanh, Webson, Raffel, Bach, Sutawika, Alyafeai, Chaffin, Stiegler, Raja, Dey, Bari, Xu, Thakker, Sharma, Szczechla, Kim, Chhablani, Nayak, Datta, Chang, Jiang, Wang, Manica, Shen, Yong, Pandey, Bawden, Wang, Neeraj, Rozen, Sharma, Santilli, Fevry, Fries, Teehan, Scao, Biderman, Gao, Wolf, and Rush}]{sanh2022multitask}
Victor Sanh, Albert Webson, Colin Raffel, Stephen Bach, Lintang Sutawika, Zaid Alyafeai, Antoine Chaffin, Arnaud Stiegler, Arun Raja, Manan Dey, M~Saiful Bari, Canwen Xu, Urmish Thakker, Shanya~Sharma Sharma, Eliza Szczechla, Taewoon Kim, Gunjan Chhablani, Nihal Nayak, Debajyoti Datta, Jonathan Chang, Mike Tian-Jian Jiang, Han Wang, Matteo Manica, Sheng Shen, Zheng~Xin Yong, Harshit Pandey, Rachel Bawden, Thomas Wang, Trishala Neeraj, Jos Rozen, Abheesht Sharma, Andrea Santilli, Thibault Fevry, Jason~Alan Fries, Ryan Teehan, Teven~Le Scao, Stella Biderman, Leo Gao, Thomas Wolf, and Alexander~M Rush. 2022.
\newblock \href {https://openreview.net/forum?id=9Vrb9D0WI4} {Multitask prompted training enables zero-shot task generalization}.
\newblock In \emph{International Conference on Learning Representations}.

\bibitem[{Satyapanich et~al.(2020)Satyapanich, Ferraro, and Finin}]{Satyapanich_Ferraro_Finin_2020}
Taneeya Satyapanich, Francis Ferraro, and Tim Finin. 2020.
\newblock \href {https://doi.org/10.1609/aaai.v34i05.6401} {Casie: Extracting cybersecurity event information from text}.
\newblock In \emph{Proceedings of the AAAI Conference on Artificial Intelligence}, volume~34, pages 8749--8757.

\bibitem[{Scheurer et~al.(2023)Scheurer, Campos, Korbak, Chan, Chen, Cho, and Perez}]{scheurer2023training}
J{\'e}r{\'e}my Scheurer, Jon~Ander Campos, Tomasz Korbak, Jun~Shern Chan, Angelica Chen, Kyunghyun Cho, and Ethan Perez. 2023.
\newblock \href {https://arxiv.org/abs/2303.16755} {Training language models with language feedback at scale}.
\newblock \emph{arXiv preprint arXiv:2303.16755}.

\bibitem[{Shin et~al.(2021)Shin, Lin, Thomson, Chen, Roy, Platanios, Pauls, Klein, Eisner, and Van~Durme}]{shin-etal-2021-constrained}
Richard Shin, Christopher Lin, Sam Thomson, Charles Chen, Subhro Roy, Emmanouil~Antonios Platanios, Adam Pauls, Dan Klein, Jason Eisner, and Benjamin Van~Durme. 2021.
\newblock \href {https://doi.org/10.18653/v1/2021.emnlp-main.608} {Constrained language models yield few-shot semantic parsers}.
\newblock In \emph{Proceedings of the 2021 Conference on Empirical Methods in Natural Language Processing}, pages 7699--7715, Online and Punta Cana, Dominican Republic. Association for Computational Linguistics.

\bibitem[{Stiennon et~al.(2020)Stiennon, Ouyang, Wu, Ziegler, Lowe, Voss, Radford, Amodei, and Christiano}]{summarization}
Nisan Stiennon, Long Ouyang, Jeffrey Wu, Daniel Ziegler, Ryan Lowe, Chelsea Voss, Alec Radford, Dario Amodei, and Paul~F Christiano. 2020.
\newblock \href {https://proceedings.neurips.cc/paper_files/paper/2020/file/1f89885d556929e98d3ef9b86448f951-Paper.pdf} {Learning to summarize with human feedback}.
\newblock In \emph{Advances in Neural Information Processing Systems}, volume~33, pages 3008--3021. Curran Associates, Inc.

\bibitem[{Sun et~al.(2024)Sun, Shaib, and Wallace}]{sun2023evaluating}
Jiuding Sun, Chantal Shaib, and Byron~C Wallace. 2024.
\newblock \href {https://openreview.net/forum?id=g9diuvxN6D} {Evaluating the zero-shot robustness of instruction-tuned language models}.
\newblock In \emph{The Twelfth International Conference on Learning Representations}.

\bibitem[{Sun et~al.(2022)Sun, Li, Pergola, Wallace, John, Greene, Kim, and He}]{PHEE}
Zhaoyue Sun, Jiazheng Li, Gabriele Pergola, Byron Wallace, Bino John, Nigel Greene, Joseph Kim, and Yulan He. 2022.
\newblock \href {https://doi.org/10.18653/v1/2022.emnlp-main.376} {{PHEE}: A dataset for pharmacovigilance event extraction from text}.
\newblock In \emph{Proceedings of the 2022 Conference on Empirical Methods in Natural Language Processing}, pages 5571--5587, Abu Dhabi, United Arab Emirates. Association for Computational Linguistics.

\bibitem[{Suzgun et~al.(2023)Suzgun, Scales, Sch{\"a}rli, Gehrmann, Tay, Chung, Chowdhery, Le, Chi, Zhou, and Wei}]{BBH}
Mirac Suzgun, Nathan Scales, Nathanael Sch{\"a}rli, Sebastian Gehrmann, Yi~Tay, Hyung~Won Chung, Aakanksha Chowdhery, Quoc Le, Ed~Chi, Denny Zhou, and Jason Wei. 2023.
\newblock \href {https://doi.org/10.18653/v1/2023.findings-acl.824} {Challenging {BIG}-bench tasks and whether chain-of-thought can solve them}.
\newblock In \emph{Findings of the Association for Computational Linguistics: ACL 2023}, pages 13003--13051, Toronto, Canada. Association for Computational Linguistics.

\bibitem[{Takanobu et~al.(2019)Takanobu, Zhang, Liu, and Huang}]{Takanobu2018AHF}
Ryuichi Takanobu, Tianyang Zhang, Jiexi Liu, and Minlie Huang. 2019.
\newblock \href {https://doi.org/10.1609/aaai.v33i01.33017072} {A hierarchical framework for relation extraction with reinforcement learning}.
\newblock \emph{Proceedings of the AAAI Conference on Artificial Intelligence}, 33(01):7072--7079.

\bibitem[{Taori et~al.(2023)Taori, Gulrajani, Zhang, Dubois, Li, Guestrin, Liang, and Hashimoto}]{alpaca}
Rohan Taori, Ishaan Gulrajani, Tianyi Zhang, Yann Dubois, Xuechen Li, Carlos Guestrin, Percy Liang, and Tatsunori~B. Hashimoto. 2023.
\newblock \href {https://github.com/tatsu-lab/stanford_alpaca} {Stanford alpaca: An instruction-following llama model}.
\newblock GitHub repository.

\bibitem[{Tedeschi et~al.(2021)Tedeschi, Maiorca, Campolungo, Cecconi, and Navigli}]{wikineuralDATASET}
Simone Tedeschi, Valentino Maiorca, Niccol{\`o} Campolungo, Francesco Cecconi, and Roberto Navigli. 2021.
\newblock \href {https://doi.org/10.18653/v1/2021.findings-emnlp.215} {{W}iki{NE}u{R}al: {C}ombined neural and knowledge-based silver data creation for multilingual {NER}}.
\newblock In \emph{Findings of the Association for Computational Linguistics: EMNLP 2021}, pages 2521--2533, Punta Cana, Dominican Republic. Association for Computational Linguistics.

\bibitem[{Tedeschi and Navigli(2022)}]{multiNERDDATASET}
Simone Tedeschi and Roberto Navigli. 2022.
\newblock \href {https://doi.org/10.18653/v1/2022.findings-naacl.60} {{M}ulti{NERD}: A multilingual, multi-genre and fine-grained dataset for named entity recognition (and disambiguation)}.
\newblock In \emph{Findings of the Association for Computational Linguistics: NAACL 2022}, pages 801--812, Seattle, United States. Association for Computational Linguistics.

\bibitem[{Tjong Kim~Sang and De~Meulder(2003)}]{CoNLL03Dataset}
Erik~F. Tjong Kim~Sang and Fien De~Meulder. 2003.
\newblock \href {https://aclanthology.org/W03-0419} {Introduction to the {C}o{NLL}-2003 shared task: Language-independent named entity recognition}.
\newblock In \emph{Proceedings of the Seventh Conference on Natural Language Learning at {HLT}-{NAACL} 2003}, pages 142--147.

\bibitem[{Touvron et~al.(2023{\natexlab{a}})Touvron, Lavril, Izacard, Martinet, Lachaux, Lacroix, Rozi{\`e}re, Goyal, Hambro, Azhar et~al.}]{LLaMA}
Hugo Touvron, Thibaut Lavril, Gautier Izacard, Xavier Martinet, Marie-Anne Lachaux, Timoth{\'e}e Lacroix, Baptiste Rozi{\`e}re, Naman Goyal, Eric Hambro, Faisal Azhar, et~al. 2023{\natexlab{a}}.
\newblock \href {https://arxiv.org/abs/2302.13971} {Llama: Open and efficient foundation language models}.
\newblock \emph{arXiv preprint arXiv:2302.13971}.

\bibitem[{Touvron et~al.(2023{\natexlab{b}})Touvron, Martin, Stone, Albert, Almahairi, Babaei, Bashlykov, Batra, Bhargava, Bhosale et~al.}]{LLaMA2}
Hugo Touvron, Louis Martin, Kevin Stone, Peter Albert, Amjad Almahairi, Yasmine Babaei, Nikolay Bashlykov, Soumya Batra, Prajjwal Bhargava, Shruti Bhosale, et~al. 2023{\natexlab{b}}.
\newblock \href {https://arxiv.org/abs/2307.09288} {Llama 2: Open foundation and fine-tuned chat models}.
\newblock \emph{arXiv preprint arXiv:2307.09288}.

\bibitem[{Ushio et~al.(2022)Ushio, Barbieri, Sousa, Neves, and Camacho-Collados}]{tweetNER7DATASET}
Asahi Ushio, Francesco Barbieri, Vitor Sousa, Leonardo Neves, and Jose Camacho-Collados. 2022.
\newblock \href {https://aclanthology.org/2022.aacl-main.25} {Named entity recognition in {T}witter: A dataset and analysis on short-term temporal shifts}.
\newblock In \emph{Proceedings of the 2nd Conference of the Asia-Pacific Chapter of the Association for Computational Linguistics and the 12th International Joint Conference on Natural Language Processing (Volume 1: Long Papers)}, pages 309--319, Online only. Association for Computational Linguistics.

\bibitem[{Walker et~al.(2006)Walker, Strassel, Medero, and Maeda}]{ace2005-annotation}
Christopher Walker, Stephanie Strassel, Julie Medero, and Kazuaki Maeda. 2006.
\newblock \href {https://doi.org/10.35111/mwxc-vh88} {Ace 2005 multilingual training corpus}.

\bibitem[{Wang et~al.(2022{\natexlab{a}})Wang, Liu, Chen, Hong, Tang, and Song}]{wang-etal-2022-deepstruct}
Chenguang Wang, Xiao Liu, Zui Chen, Haoyun Hong, Jie Tang, and Dawn Song. 2022{\natexlab{a}}.
\newblock \href {https://doi.org/10.18653/v1/2022.findings-acl.67} {{D}eep{S}truct: Pretraining of language models for structure prediction}.
\newblock In \emph{Findings of the Association for Computational Linguistics: ACL 2022}, pages 803--823, Dublin, Ireland. Association for Computational Linguistics.

\bibitem[{Wang et~al.(2023{\natexlab{a}})Wang, Zhou, Zu, Xia, Chen, Zhang, Zheng, Ye, Zhang, Gui et~al.}]{wang2023instructuie}
Xiao Wang, Weikang Zhou, Can Zu, Han Xia, Tianze Chen, Yuansen Zhang, Rui Zheng, Junjie Ye, Qi~Zhang, Tao Gui, et~al. 2023{\natexlab{a}}.
\newblock \href {https://arxiv.org/abs/2304.08085} {Instructuie: Multi-task instruction tuning for unified information extraction}.
\newblock \emph{arXiv preprint arXiv:2304.08085}.

\bibitem[{Wang et~al.(2023{\natexlab{b}})Wang, Kordi, Mishra, Liu, Smith, Khashabi, and Hajishirzi}]{wang-etal-2023-self-instruct}
Yizhong Wang, Yeganeh Kordi, Swaroop Mishra, Alisa Liu, Noah~A. Smith, Daniel Khashabi, and Hannaneh Hajishirzi. 2023{\natexlab{b}}.
\newblock \href {https://doi.org/10.18653/v1/2023.acl-long.754} {Self-instruct: Aligning language models with self-generated instructions}.
\newblock In \emph{Proceedings of the 61st Annual Meeting of the Association for Computational Linguistics (Volume 1: Long Papers)}, pages 13484--13508, Toronto, Canada. Association for Computational Linguistics.

\bibitem[{Wang et~al.(2022{\natexlab{b}})Wang, Mishra, Alipoormolabashi, Kordi, Mirzaei, Naik, Ashok, Dhanasekaran, Arunkumar, Stap, Pathak, Karamanolakis, Lai, Purohit, Mondal, Anderson, Kuznia, Doshi, Pal, Patel, Moradshahi, Parmar, Purohit, Varshney, Kaza, Verma, Puri, Karia, Doshi, Sampat, Mishra, Reddy~A, Patro, Dixit, and Shen}]{wang-etal-2022-super}
Yizhong Wang, Swaroop Mishra, Pegah Alipoormolabashi, Yeganeh Kordi, Amirreza Mirzaei, Atharva Naik, Arjun Ashok, Arut~Selvan Dhanasekaran, Anjana Arunkumar, David Stap, Eshaan Pathak, Giannis Karamanolakis, Haizhi Lai, Ishan Purohit, Ishani Mondal, Jacob Anderson, Kirby Kuznia, Krima Doshi, Kuntal~Kumar Pal, Maitreya Patel, Mehrad Moradshahi, Mihir Parmar, Mirali Purohit, Neeraj Varshney, Phani~Rohitha Kaza, Pulkit Verma, Ravsehaj~Singh Puri, Rushang Karia, Savan Doshi, Shailaja~Keyur Sampat, Siddhartha Mishra, Sujan Reddy~A, Sumanta Patro, Tanay Dixit, and Xudong Shen. 2022{\natexlab{b}}.
\newblock \href {https://doi.org/10.18653/v1/2022.emnlp-main.340} {Super-{N}atural{I}nstructions: Generalization via declarative instructions on 1600+ {NLP} tasks}.
\newblock In \emph{Proceedings of the 2022 Conference on Empirical Methods in Natural Language Processing}, pages 5085--5109, Abu Dhabi, United Arab Emirates. Association for Computational Linguistics.

\bibitem[{Wei et~al.(2022)Wei, Bosma, Zhao, Guu, Yu, Lester, Du, Dai, and Le}]{wei2022finetuned}
Jason Wei, Maarten Bosma, Vincent Zhao, Kelvin Guu, Adams~Wei Yu, Brian Lester, Nan Du, Andrew~M. Dai, and Quoc~V Le. 2022.
\newblock \href {https://openreview.net/forum?id=gEZrGCozdqR} {Finetuned language models are zero-shot learners}.
\newblock In \emph{International Conference on Learning Representations}.

\bibitem[{Wei et~al.(2023)Wei, Hou, Lampinen, Chen, Huang, Tay, Chen, Lu, Zhou, Ma, and Le}]{wei2023symbol}
Jerry Wei, Le~Hou, Andrew Lampinen, Xiangning Chen, Da~Huang, Yi~Tay, Xinyun Chen, Yifeng Lu, Denny Zhou, Tengyu Ma, and Quoc Le. 2023.
\newblock \href {https://doi.org/10.18653/v1/2023.emnlp-main.61} {Symbol tuning improves in-context learning in language models}.
\newblock In \emph{Proceedings of the 2023 Conference on Empirical Methods in Natural Language Processing}, pages 968--979, Singapore. Association for Computational Linguistics.

\bibitem[{Xu et~al.(2024)Xu, Sun, Zheng, Geng, Zhao, Feng, Tao, Lin, and Jiang}]{xu2023wizardlm}
Can Xu, Qingfeng Sun, Kai Zheng, Xiubo Geng, Pu~Zhao, Jiazhan Feng, Chongyang Tao, Qingwei Lin, and Daxin Jiang. 2024.
\newblock \href {https://openreview.net/forum?id=CfXh93NDgH} {Wizard{LM}: Empowering large pre-trained language models to follow complex instructions}.
\newblock In \emph{The Twelfth International Conference on Learning Representations}.

\bibitem[{Ye et~al.(2024)Ye, Hwang, Yang, Yun, Kim, and Seo}]{ye2023context}
Seonghyeon Ye, Hyeonbin Hwang, Sohee Yang, Hyeongu Yun, Yireun Kim, and Minjoon Seo. 2024.
\newblock Investigating the effectiveness of task-agnostic prefix prompt for instruction following.
\newblock In \emph{Proceedings of the AAAI Conference on Artificial Intelligence}, volume~38, pages 19386--19394.

\bibitem[{Yin et~al.(2023)Yin, Liu, Yin, Zhong, Bansal, Han, and Chang}]{yin-etal-2023-dynosaur}
Da~Yin, Xiao Liu, Fan Yin, Ming Zhong, Hritik Bansal, Jiawei Han, and Kai-Wei Chang. 2023.
\newblock \href {https://doi.org/10.18653/v1/2023.emnlp-main.245} {Dynosaur: A dynamic growth paradigm for instruction-tuning data curation}.
\newblock In \emph{Proceedings of the 2023 Conference on Empirical Methods in Natural Language Processing}, pages 4031--4047, Singapore. Association for Computational Linguistics.

\bibitem[{Yu et~al.(2018)Yu, Zhang, Yang, Yasunaga, Wang, Li, Ma, Li, Yao, Roman, Zhang, and Radev}]{yu-etal-2018-spider}
Tao Yu, Rui Zhang, Kai Yang, Michihiro Yasunaga, Dongxu Wang, Zifan Li, James Ma, Irene Li, Qingning Yao, Shanelle Roman, Zilin Zhang, and Dragomir Radev. 2018.
\newblock \href {https://doi.org/10.18653/v1/D18-1425} {{S}pider: A large-scale human-labeled dataset for complex and cross-domain semantic parsing and text-to-{SQL} task}.
\newblock In \emph{Proceedings of the 2018 Conference on Empirical Methods in Natural Language Processing}, pages 3911--3921, Brussels, Belgium. Association for Computational Linguistics.

\bibitem[{Zhang and Wang(2015)}]{kbp37DATASET}
Dongxu Zhang and Dong Wang. 2015.
\newblock \href {https://arxiv.org/abs/1508.01006} {Relation classification via recurrent neural network}.
\newblock \emph{arXiv preprint arXiv:1508.01006}.

\bibitem[{Zhang et~al.(2022)Zhang, Roller, Goyal, Artetxe, Chen, Chen, Dewan, Diab, Li, Lin et~al.}]{zhang2022opt}
Susan Zhang, Stephen Roller, Naman Goyal, Mikel Artetxe, Moya Chen, Shuohui Chen, Christopher Dewan, Mona Diab, Xian Li, Xi~Victoria Lin, et~al. 2022.
\newblock \href {https://arxiv.org/abs/2205.01068} {Opt: Open pre-trained transformer language models}.
\newblock \emph{arXiv preprint arXiv:2205.01068}.

\bibitem[{Zheng et~al.(2024)Zheng, Zhou, Meng, Zhou, and Huang}]{zheng2024large}
Chujie Zheng, Hao Zhou, Fandong Meng, Jie Zhou, and Minlie Huang. 2024.
\newblock \href {https://openreview.net/forum?id=shr9PXz7T0} {Large language models are not robust multiple choice selectors}.
\newblock In \emph{The Twelfth International Conference on Learning Representations}.

\bibitem[{Zhou et~al.(2023)Zhou, Liu, Xu, Iyer, Sun, Mao, Ma, Efrat, Yu, YU, Zhang, Ghosh, Lewis, Zettlemoyer, and Levy}]{zhou2023lima}
Chunting Zhou, Pengfei Liu, Puxin Xu, Srini Iyer, Jiao Sun, Yuning Mao, Xuezhe Ma, Avia Efrat, Ping Yu, LILI YU, Susan Zhang, Gargi Ghosh, Mike Lewis, Luke Zettlemoyer, and Omer Levy. 2023.
\newblock \href {https://openreview.net/forum?id=KBMOKmX2he} {{LIMA}: Less is more for alignment}.
\newblock In \emph{Thirty-seventh Conference on Neural Information Processing Systems}.

\end{thebibliography}
\end{document}